\PassOptionsToPackage{table,xcdraw,dvipsnames}{xcolor}
\documentclass{article} 
\usepackage{natbib}
\usepackage[table,xcdraw,dvipsnames]{xcolor}
\usepackage{arxiv_iclr2025_conference,times}

\usepackage{amsmath,amsfonts,bm}

\def\eqref#1{equation~\ref{#1}}

\def\1{\bm{1}}

\DeclareMathAlphabet{\mathsfit}{\encodingdefault}{\sfdefault}{m}{sl}
\SetMathAlphabet{\mathsfit}{bold}{\encodingdefault}{\sfdefault}{bx}{n}

\usepackage{hyperref}
\usepackage{url}            
\usepackage{booktabs}       
\usepackage{nicefrac}       
\usepackage{microtype}      
\usepackage{xcolor}         
\usepackage[table]{xcolor}
\usepackage{algorithm}
\usepackage{algorithmic}

\usepackage{amsmath}  
\usepackage{comment}  
\usepackage{graphicx}
\usepackage{subcaption}

\usepackage{threeparttable}

\usepackage{amsfonts} 
\usepackage{enumitem}

\usepackage{adjustbox}
\usepackage{tabularray}
\usepackage{nicematrix}
\usepackage{pifont}
\newcommand{\cmark}{\ding{51}}
\newcommand{\xmark}{\ding{55}}
\usepackage{multirow}

\usepackage{wrapfig}

\usepackage{tablefootnote}

\usepackage{titlesec} 
\titlespacing*{\section} {0pt}{4pt}{2pt}
\newcommand{\first}[1]{\textbf{\textcolor{red}{#1}}}
\newcommand{\second}[1]{\underline{\textcolor{blue}{#1}}}
\newcommand{\third}[1]{\textcolor{black}{#1}}
\newcommand{\fourth}[1]{\textcolor{black}{#1}}

\titlespacing*{\paragraph} {0pt}{2pt}{4pt}
\title{Sequential Order-Robust Mamba for \\ Time Series Forecasting}
\vspace{-14pt}
\author{Seunghan Lee\thanks{Equal contribution.}, Juri Hong\footnotemark[1], Kibok Lee\thanks{Equal advising.}, Taeyoung Park\footnotemark[2]
\\
Department of Statistics and Data Science, Yonsei University\\  
\texttt{\{seunghan9613,jurih,kibok,tpark\}@yonsei.ac.kr} \\
}

\iclrfinalcopy 
\begin{document}
\maketitle
\vspace{-17pt}
\begin{abstract}
Mamba has recently emerged as a promising alternative to Transformers, offering near-linear complexity in processing sequential data.
However, while channels in time series (TS) data have no specific order in general, recent studies have adopted Mamba to capture channel dependencies (CD) in TS, introducing a 
\textit{sequential order bias}.
To address this issue, we propose SOR-Mamba, a TS forecasting method that 1) incorporates a regularization strategy to minimize the discrepancy between two embedding vectors generated from data with reversed channel orders, thereby enhancing robustness to channel order, and 2) eliminates the 1D-convolution originally designed to capture local information in sequential data.
Furthermore, we introduce channel correlation modeling (CCM), a pretraining task aimed at preserving correlations between channels from the data space to the latent space
in order to enhance the ability to capture CD.
Extensive experiments demonstrate the efficacy of the proposed method across standard and transfer learning scenarios.
Code is available at \url{https://github.com/seunghan96/SOR-Mamba}.
\end{abstract}

\section{Introduction}
Time series (TS) forecasting is prevalent in various fields, including 
weather~\citep{angryk2020multivariate}, traffic~\citep{cirstea2022towards}, and energy~\citep{dudek2021hybrid}. 
While Transformers~\citep{vaswani2017attention} have been widely employed for this task due to their ability to capture long-term dependencies in 
sequences~\citep{wen2022transformers}, their quadratic computational complexity causes substantial computational overhead, limiting their practicality in real-world applications.
Several attempts have been made to reduce the complexity of Transformers~\citep{zhang2023crossformer,zhou2022fedformer};
however, they often result in performance degradations~\citep{wang2024mamba}.

To tackle the computational challenges of
Transformers,
alternatives
such as state-space models (SSMs)~\citep{gu2021efficiently} 
have been considered, employing convolutional operations
to process sequences with linear complexity.
Recently, Mamba~\citep{gu2023mamba} enhanced SSMs by incorporating a selective mechanism 
to prioritize important information efficiently.
Due to its strong balance between performance and computational efficiency~\citep{wang2024mamba}, Mamba has been widely adopted 
across various domains~\citep{zhu2024vision,schiff2024caduceus}. 
In the TS domain, Mamba is utilized to capture temporal dependencies (TD) by processing input TS along the \textit{temporal dimension} \citep{ahamed2024timemachine}, channel dependencies (CD) along the \textit{channel dimension} \citep{wang2024mamba}, or both~\citep{cai2024mambats}.
In this paper, we focus on Mamba capturing CD, in line with the recent work~\citep{liu2023itransformer} that advocates for the use of complex attention mechanisms for CD while employing simple multi-layer perceptrons (MLPs) for TD.

\newlength{\defaultcolumnsep}
\newlength{\defaultintextsep}
\newlength{\defaultabovecaptionskip}
\newlength{\defaultbelowcaptionskip}
\setlength{\intextsep}{0.1pt} 
\setlength{\abovecaptionskip}{3.3pt}
\begin{wrapfigure}{r}{0.44\textwidth}
  \centering
  \vspace{-5pt}
  \includegraphics[width=0.44\textwidth]{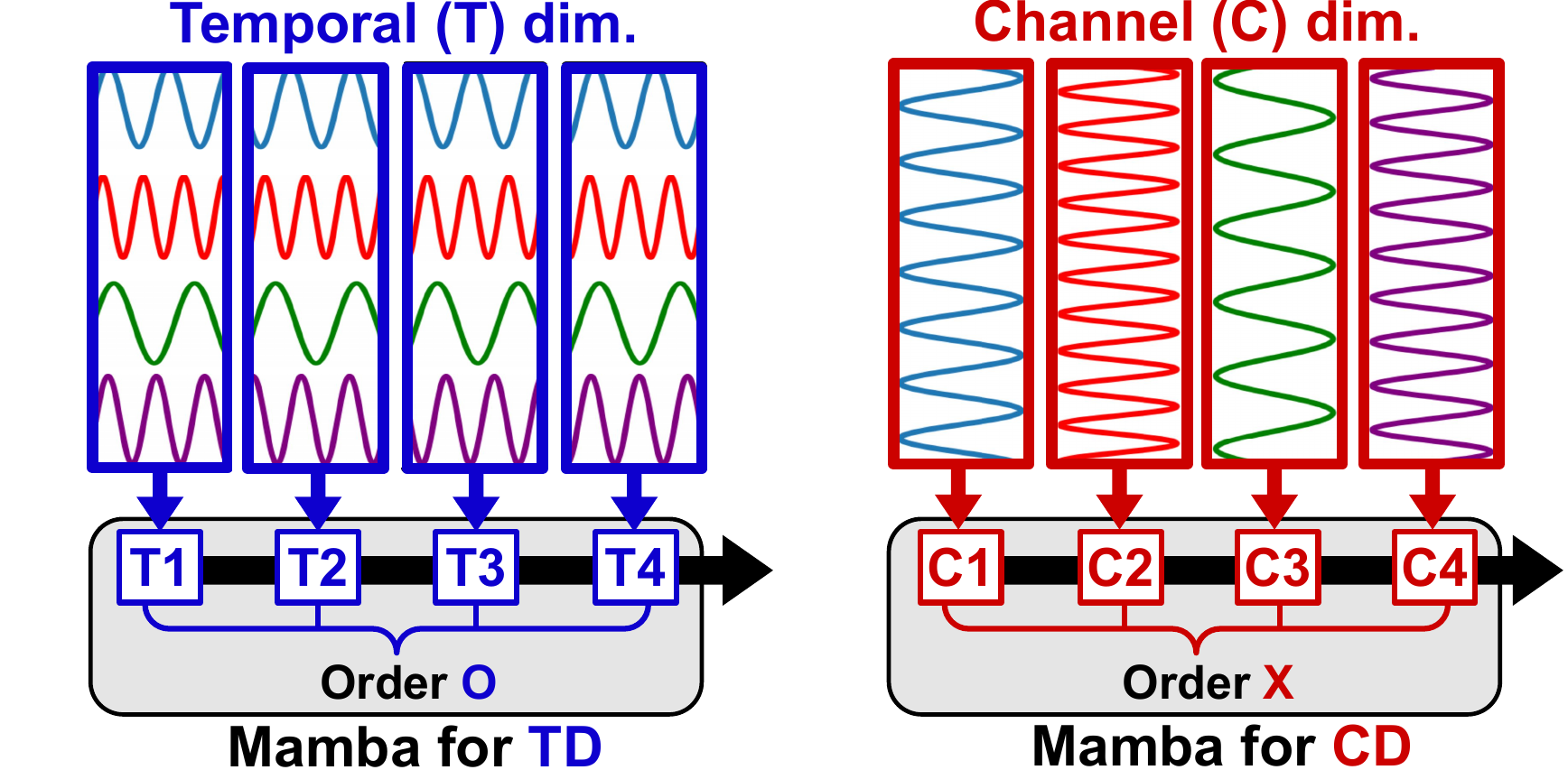} 
  \caption{Capturing CD with Mamba, which has a sequential order bias, 
  is challenging
  as channels lack an inherent sequential order.}
  \label{fig:motivation}
\end{wrapfigure}
However, applying Mamba to capture CD is challenging as channels lack an inherent sequential order, whereas Mamba is originally designed for sequential inputs (i.e., Mamba contains a \textit{sequential order bias}), as shown in Figure~\ref{fig:motivation}.
To address this issue, 
previous works have employed the bidirectional Mamba to capture 
CD~\citep{wang2024mamba, liang2024bi}, 
where two unidirectional Mambas with different parameters 
capture CD from a certain channel order and its reversed order.
However, these methods are inefficient due to the need for two models. 
Another approach involves permuting a channel order during training~\citep{cai2024mambats} to enhance robustness to the order, 
while requiring
an additional procedure to determine the optimal order for inference.

\setlength{\intextsep}{1.2pt} 
\setlength{\columnsep}{4pt}
\setlength{\intextsep}{0.1pt} 
\begin{wraptable}{r}{0.49\textwidth}
\vspace{-4pt}
\centering
\begin{adjustbox}{max width=0.49\textwidth}
\begin{NiceTabular}{c|cccc}
\toprule
\multirow{2.5}{*}{\shortstack{ECL dataset\\(Metric: MSE)}}& \multicolumn{4}{c}{Horizon } \\
\cmidrule(lr){2-5}
& 96 & 192 & 336 & 720 \\
\midrule
\multicolumn{1}{c}{Bidirectional} & \first{0.139} & 0.165 & \first{0.177} & 0.214 \\
\midrule
 \textcircled{1} Uni ($1 \rightarrow C$) & 0.143 & \first{0.162} & 0.179 &0.234 \\
\textcircled{2} Uni ($C \rightarrow 1$) & 0.141 & 0.168 & 0.179 & \first{0.210} \\
\midrule
\multicolumn{1}{c}{ (\textcircled{1} - \textcircled{2}) / \textcircled{1}} & \cellcolor{gray!20} \textbf{-1.6\%} & \cellcolor{gray!20} \textbf{+3.8\%} & \cellcolor{gray!20} \textbf{+0.2\%} & \cellcolor{gray!20} \textbf{-10.3\%}\\
\bottomrule
\end{NiceTabular}
\end{adjustbox}
\caption{
Bidirectional Mamba may not achieve the best performance, and the performance of the unidirectional Mamba varies by the channel order.}
\label{tbl:motivation2}
\end{wraptable}
Furthermore, Table~\ref{tbl:motivation2} shows the performance of the TS forecasting task 
using the bidirectional Mamba and two unidirectional Mambas with reversed channel orders, suggesting that the bidirectional Mamba~\citep{wang2024mamba} may not be effective in handling the sequential order bias.
The table indicates that 1) the bidirectional Mamba does not always achieve 
the best performance, 
and 2) the performance of the unidirectional Mamba varies depending on the channel order.

To this end, we introduce \textbf{S}equential \textbf{O}rder-\textbf{R}obust \textbf{Mamba} for TS forecasting (\textit{SOR-Mamba}), 
a TS forecasting method that 
handles the sequential order bias by
1) incorporating
a regularization strategy
to minimize the distance between two embedding vectors generated 
from data 
with reversed channel orders
to enhance robustness to the order,
and 2) removing the 1D-convolution (1D-conv) originally designed to capture local information in sequential inputs.
Additionally, we propose \textbf{C}hannel \textbf{C}orrelation \textbf{M}odeling (\textit{CCM}),
a pretraining task 
aimed at improving the model's ability to capture CD by preserving the correlation between channels from the data space to the latent space.
The main contributions of this work are summarized as follows:
\setlist[itemize]{leftmargin=.3cm}
\begin{itemize}
    \item We propose SOR-Mamba, 
    a TS forecasting method that handles the sequential order bias
    by 1)~regularizing the unidirectional Mamba to minimize the distance between two embedding vectors 
    generated 
    from data 
    with reversed channel orders
    for robustness to channel order
    and 2)~removing the 1D-conv from the original Mamba block, as channels lack an inherent sequential order.
    \item We introduce CCM, a novel pretraining task that preserves the correlation between channels from the data space to the latent space, thereby enhancing the model's ability to capture CD.
    \item We conduct extensive experiments 
    with 13 datasets
    in both standard and transfer learning settings,
    demonstrating that our method achieves state-of-the-art (SOTA) performance with greater efficiency compared to previous SOTA methods by utilizing the unidirectional Mamba.
\end{itemize}

\section{Related Works}
\textbf{TS forecasting with Transformer.}
Transformers~\citep{vaswani2017attention} are commonly employed for long-term TS forecasting (LTSF) tasks 
due to
their ability to handle long-range dependencies through attention mechanisms. 
However, their quadratic complexity has led to the development of various methods aimed at improving efficiency, 
such as 
modifying the
Transformer architecture~\citep{zhang2023crossformer,zhou2022fedformer},
patchifying the TS~\citep{nie2022time} or using MLP-based models~\citep{chen2023tsmixer,zeng2023transformers}. 
While MLP-based models offer simpler structures and reduced complexity compared to Transformers, 
they tend to be less effective at capturing global dependencies~\citep{wang2024mamba}.
Recently, iTransformer~\citep{liu2023itransformer} inverts the conventional Transformer framework in the TS domain by treating each channel as a token rather than each patch, shifting the focus from capturing TD to CD. This framework has led to significant performance gains and has become widely adopted as the backbone for TS models~\citep{liutimer,dong2024timesiam}.

\textbf{State-space models.}
To overcome the limitations of Transformer-based models, 
state-space models have been integrated with deep learning to tackle the challenge of long-range dependencies~\citep{rangapuram2018deep, zhang2023effectively,zhou2023deep}. 
However, these methods are unable to adapt their internal parameters to varying inputs, which limits their performance.
Recently, Mamba~\citep{gu2023mamba} introduces a selective scan mechanism that efficiently filters specific inputs and captures long-range context 
by incorporating time-varying parameters into the SSM.
Due to its linear-time efficiency for modeling long sequences,
it has been widely adopted in various domains, including computer vision~\citep{ma2024u, huang2024localmamba,zhu2024vision} and natural language processing~\citep{pioro2024moe, anthony2024blackmamba, he2024densemamba}.

\textbf{TS forecasting with Mamba.}
Due to its balance between performance and computational efficiency, 
Mamba has also been applied in the TS domain.
TimeMachine~\citep{ahamed2024timemachine} utilizes multi-scale quadruple-Mamba to capture either TD alone or both TD and CD, 
with its architecture relying on the statistics of the dataset.
CMamba~\citep{zeng2024c} captures TD with patch-wise Mamba and CD with an MLP.
FMamba~\citep{ma2024fmamba} integrates fast-attention with Mamba to capture CD,
and
SST~\citep{xusst} captures global and local patterns in TS with Mamba and Transformer, respectively.
S-Mamba~\citep{wang2024mamba}, Bi-Mamba+~\citep{liang2024bi}, and SAMBA~\citep{weng2024simplified}, designed to capture CD in TS,
use bidirectional scanning with the bidirectional Mamba to address the sequential order bias, although they are limited by the need for two models.
MambaTS~\citep{cai2024mambats} introduces variable permutation training, which shuffles the channel order during the training stage to handle the sequential order bias. 
However, it is limited by the need for an additional procedure to determine the optimal scan order for the inference stage.

\section{Preliminaries}
\textbf{Problem definition.} 
This paper addresses the multivariate TS forecasting task, 
where the model 
uses a lookback window $\mathbf{x}=(\mathbf{x}_1, \mathbf{x}_2, \cdots, \mathbf{x}_L)$
to predict future values $\mathbf{y}=(\mathbf{x}_{L+1}, \cdots, \mathbf{x}_{L+H})$ with $\mathbf{x}_i \in \mathbb{R}^C$ representing the values at each time step.
Here, $L$, $H$, and $C$ denote the size of the lookback window, the forecast horizon, and the number of channels, respectively.

\textbf{State-space models.} 
SSM
transforms the
continuous input signals $x(t)$ into corresponding outputs $y(t)$ via a state representation $h(t)$. 
This state space represents how the state evolves over time, 
which can be expressed using ordinary differential equations as follows:
\begin{equation}
    \begin{aligned}
        h^\prime(t) & =\boldsymbol{A}h(t)+\boldsymbol{B}x(t), \\
        y(t) & =\boldsymbol{C}h(t)+\boldsymbol{D}x(t),
    \end{aligned}
\end{equation}
where $h^\prime(t)=\frac{dh(t)}{dt}$, and $\boldsymbol{A}, \boldsymbol{B}, \boldsymbol{C}$, and $\boldsymbol{D}$ are learnable parameters of the SSMs.

Due to the continuous nature of SSMs, 
discretization is commonly used to approximate continuous-time representations into discrete-time representations by sampling input signals at fixed intervals.
This results in the discrete-time SSMs being represented as:
\begin{equation}
	\begin{aligned}
		h_k & =\overline{\boldsymbol{A}} h_{k-1}+\overline{\boldsymbol{B}} x_k, \\
		y_k & ={\boldsymbol{C}} h_k+{\boldsymbol{D}} x_k,
	\end{aligned}
\end{equation}
where $h_k$ and $x_k$
are the state vector and input vector at time $k$, respectively,
and $\overline{\boldsymbol{A}}=\exp (\Delta \boldsymbol{A})$ and $\overline{\boldsymbol{B}}=(\Delta \boldsymbol{A})^{-1}(\exp (\Delta \boldsymbol{A})-\boldsymbol{I}) \cdot \Delta \boldsymbol{B}$ are the discrete-time matrices obtained from the 
$A$ and $B$. 

Recently, Mamba introduces selective SSMs
that enables the model to capture contextual information in long sequences
using time-varying parameters~\citep{gu2023mamba}.
Its near-linear complexity makes it an efficient alternative to the quadratic complexity of the attention mechanism in Transformers.

\begin{figure*}[t]
\vspace{-25pt}
\centering
\includegraphics[width=.999\textwidth]{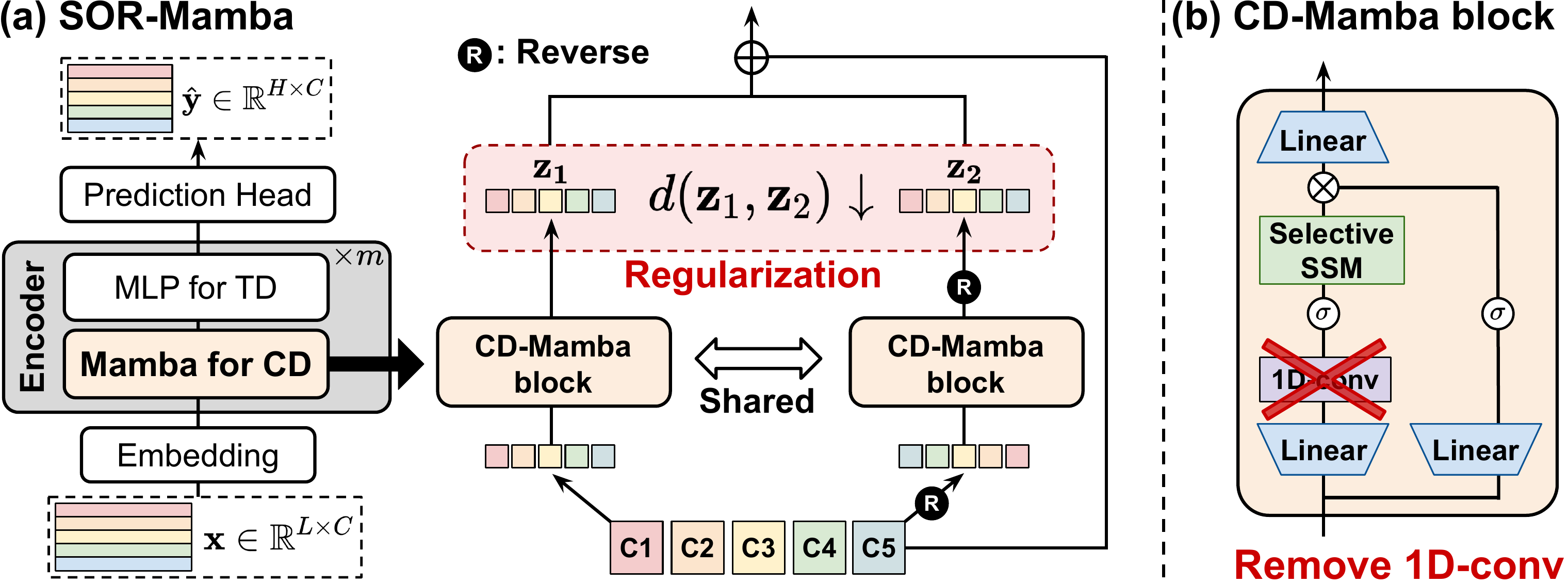} 
\caption{\textbf{Overall framework of SOR-Mamba.} 
(a) shows the architecture of SOR-Mamba,
where the CD-Mamba block is regularized to minimize the distance between two vectors derived from reversed channel orders. 
(b) shows the CD-Mamba block, 
where the 1D-conv from the Mamba block is removed, as channels do not have a sequential order, which is further explained in Appendix~\ref{sec:1d_conv_remove}.
}
\label{fig:main1}
\vspace{-14pt}
\end{figure*}
\section{Methodology}
\label{sec:pi_task}
In this paper, we introduce SOR-Mamba, 
a TS forecasting method
designed to address the sequential order bias 
by 1) regularizing Mamba
to minimize the distance between two embedding vectors generated from data with reversed channel orders
and 
2) removing the 1D-conv from the original Mamba block.
The overall framework of SOR-Mamba is illustrated in Figure~\ref{fig:main1},
which consists of four components: 
the embedding layer for tokenization, 
Mamba for capturing CD, 
MLP for capturing TD, 
and the prediction head for predicting the future output.

Furthermore, we introduce a novel pretraining task, CCM, where the model is pretrained to preserve the correlation between channels from the data space to the latent space, aligning with the recent TS models that focus on capturing CD over TD.
The overall framework of CCM is illustrated in Figure~\ref{fig:main2}.

\subsection{Architecture of SOR-Mamba} 
\textbf{1) Embedding layer.}
To tokenize the TS in a channel-wise manner, 
we use an embedding layer that treats each channel as a token, 
following the approach in iTransformer \citep{liu2023itransformer}. 
Specifically, we transform $\mathbf{x} \in \mathbb{R}^{L \times C}$ into $\mathbf{z} \in \mathbb{R}^{C \times D}$ using a single linear layer.

\textbf{2) Mamba for CD.} 
The original Mamba block combines the H3 block~\citep{fu2022hungry} with a gated MLP, where the H3 block incorporates a 1D-conv before the SSM layer to capture local information from adjacent steps.
However, since channels in TS do not possess any inherent sequential order, we find this convolution unnecessary for capturing CD.
Accordingly, 
we remove the convolution from the original Mamba block, 
resulting in the proposed \textit{CD-Mamba block}, as illustrated in Figure~\ref{fig:main1}(b).
Note that this differs from the previous work~\citep{cai2024mambats} which replaces the 1D-conv with a dropout in the Mamba block, as it is designed to capture TD.
Using the CD-Mamba block, we obtain $\mathbf{z}_1$ and $\mathbf{z}_2$,
which are two embedding vectors
with reversed channel orders 
that are employed for regularization to address the sequential order bias.
These vectors are then added element-wise and combined with a residual connection from $\mathbf{z}$.
Further analysis regarding the removal of the 1D-conv can be found in Table~\ref{tbl:1dremove}.

\begin{wrapfigure}{R}{0.495\textwidth}
    \vspace{-8pt}
    \begin{minipage}{0.495\textwidth}
      \begin{algorithm}[H]
        \vspace{2pt}
        \caption{The procedure of SOR-Mamba}
        \small
        \label{alg:SOR-Mamba}
        \textbf{Input}: $\mathbf{X} = [\mathbf{X}_1, \ldots ,\mathbf{X}_L] : (B,L,C)$ \\
        \textbf{Output}: $\hat{\mathbf{Y}} = [\hat{\mathbf{X}}_{L+1},  \ldots, \hat{\mathbf{X}}_{L+H}]: (B,H,C)$ 
        
        \begin{algorithmic}[1] 
        \STATE $\mathbf{Z}:(B,C,D) \leftarrow \text{Linear} (\mathbf{X}^\top)$ 
        \FOR{$m$ in \text{layers}}
        \STATE $\mathbf{Z}_1:(B,C,D) \leftarrow \text{CD-Mamba} (\mathbf{Z}) $
        \STATE $\mathbf{Z}_2:(B,C,D) \leftarrow \text{CD-Mamba} (\mathbf{Z}^{\star})^{\star}$, 
        \\ $\quad \quad \quad \quad \quad \quad \quad \quad $ where $\mathbf{Z}^{\star} = \mathbf{Z}[:,::-1,:]$
        \STATE $\mathbf{Z}:(B,C,D) \leftarrow (\mathbf{Z}_1 + \mathbf{Z}_2) + \mathbf{Z}$
        \STATE $\mathbf{Z}:(B,C,D) \leftarrow \text{LN}(\text{MLP}(\text{LN}(\mathbf{Z})))$ 
        \ENDFOR
        \STATE $\hat{\mathbf{Y}}:(B,H,C) \leftarrow \text{Linear}(\mathbf{Z})^\top$
        \end{algorithmic}
        \end{algorithm}
    \end{minipage}
    \vspace{-14pt}
  \end{wrapfigure}
\textbf{3) MLP for TD.}
To capture TD in TS, 
we apply an MLP
to the 
output tokens of the CD-Mamba block. 
To enhance training stability, we apply layer normalization (LN) to standardize the tokens both before and after the MLP.

\textbf{4) Prediction head.}
To predict the future output, we employ a linear prediction head to the output tokens of MLP,
resulting in $\hat{\mathbf{y}} \in \mathbb{R}^{H \times C}$.
The procedure of SOR-Mamba is described in Algorithm~\ref{alg:SOR-Mamba}, where $\mathbf{Z}^{\star}$ represents $\mathbf{Z}$ with its channel order reversed.

\subsection{Regularization with CD-Mamba Block} 
To address the sequential order bias, 
SOR-Mamba regularizes the CD-Mamba block 
to minimize
the distance between two embedding vectors generated from data with reversed channel orders.
The regularization term is defined as follows:
\begin{equation}
L_{\text{reg}}(\mathbf{z})=d\left( \mathbf{z}_1,\mathbf{z}_2 \right),
\end{equation}
where $d$ is a distance metric,
and $\mathbf{z}_1$ and $\mathbf{z}_2$ are the embedding vectors obtained from the CD-Mamba block
using $\mathbf{z}$
with its channel order reversed, as described in Algorithm~\ref{alg:SOR-Mamba}.
For $d$, we use the mean squared error (MSE) in the experiments, where the robustness to the choice of 
$d$ can be found in Appendix~\ref{sec:robust_distance}.
The proposed regularization term is then added to the forecasting loss 
($L_{\text{fcst}}$)
with a contribution of $\lambda$, resulting in:
\begin{equation}
L(\mathbf{x},\mathbf{y})=L_{\text{fcst}}(\mathbf{x},\mathbf{y})+\lambda \cdot \sum_{i=1}^{m} L_{\text{reg}}(\mathbf{z}^{(i)}),
\end{equation}
where $\mathbf{z}^{(i)}$ is 
$\mathbf{z}$ at the $i$-th layer, and $m$ is the number of encoder layers.
By incorporating the regularization strategy into the unidirectional Mamba,
we achieve better performance and efficiency 
compared to S-Mamba~\citep{wang2024mamba}, which employs the bidirectional Mamba, as shown in Table~\ref{tbl:ablation}.
Additionally, we find that regularization also benefits the bidirectional Mamba, which handles the sequential order bias through bidirectional scanning,
as shown in Table~\ref{tbl:reg}. 
Further analysis regarding the robustness to $\lambda$
is discussed in Appendix~\ref{sec:robust_lambda}.

\subsection{Channel Correlation Modeling}
Previous pretraining tasks for TS have primarily focused on TD, such as masked modeling~\citep{zerveas2021transformer} and reconstruction~\citep{lee2023learning}, to pretrain an encoder.
However, we argue for the necessity of a new task 
that emphasizes CD over TD 
to align with recent TS models that focus on capturing CD 
with complex model architectures~\citep{liu2023itransformer,wang2024mamba}.
To this end, 
we propose 
CCM, 
which aims to preserve the (Pearson) correlation between channels from the data space to the latent space,
as correlation is a simple yet effective way to measure channel relationships and has been utilized in prior studies to analyze CD~\citep{yang2024vcformer,zhao2024rethinking}. 

\setlength{\columnsep}{10pt}
\setlength{\intextsep}{0.1pt} 
\begin{wrapfigure}{r}{0.422\textwidth}
  \centering
  \includegraphics[width=0.422\textwidth]{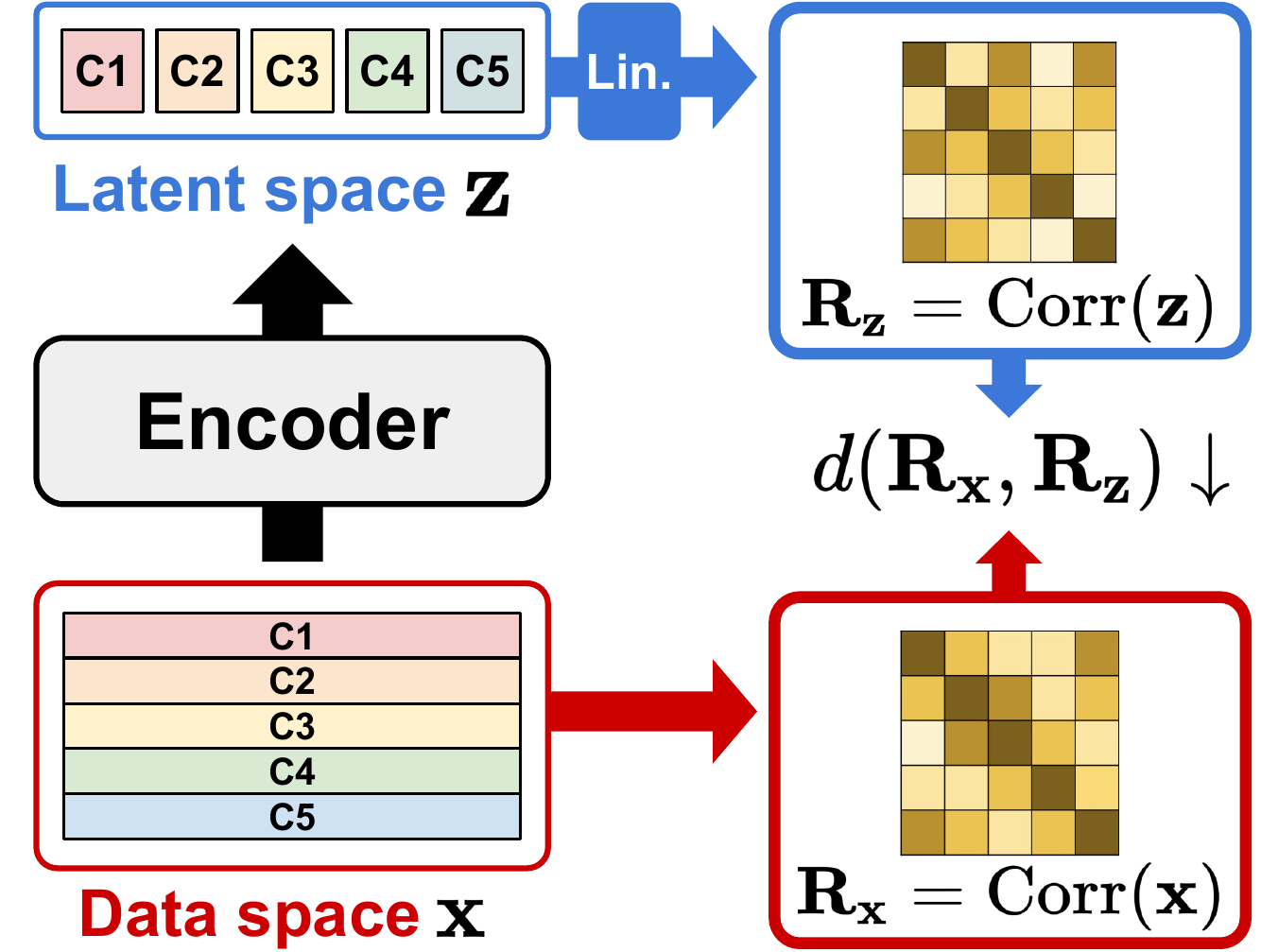}
    \caption{Channel correlation modeling.}
  \label{fig:main2}
  \vspace{10pt}
\end{wrapfigure}
For CCM, we calculate the correlation matrices between the input token on the data space
and the output token after the additional linear projection layer on the latent space,
as shown in Figure~\ref{fig:main2}.
The loss function for CCM, defined as the distance between these two matrices, can be expressed as:
\begin{equation}
L_{\text{CCM}}(\mathbf{x})=d\left( \mathbf{R_x}, \mathbf{R_z} \right),
\end{equation}
where $\mathbf{R_x}$ and $\mathbf{R_z}$ are the correlation matrices in the data space and the latent space, respectively.
We find that CCM is more effective than masked modeling and reconstruction across diverse datasets with varying numbers of channels,
as demonstrated in Table~\ref{tbl:pretraining_tasks}.
Additionally, its performance remains robust regardless of the choice of 
$d$, as discussed in Appendix~\ref{sec:robust_distance}.

\section{Experiments}
\subsection{Experimental Settings}
\textbf{Tasks and evaluation metrics.}
We demonstrate the effectiveness of SOR-Mamba on TS forecasting tasks 
with 13 datasets
under standard and transfer learning settings.
For evaluation, we follow the standard self-supervised learning (SSL) framework, which involves pretraining and fine-tuning (FT) or linear probing (LP) on the same dataset. 
Additionally, we consider in-domain and cross-domain transfer learning settings, with the domains defined in the previous work~\citep{dong2023simmtm}.
For evaluation metrics, we employ mean squared error (MSE) and mean absolute error (MAE).

\textbf{Datasets.}
For the forecasting tasks, we use 13 datasets: 
four ETT datasets (ETTh1, ETTh2, ETTm1, ETTm2)~\citep{zhou2021informer}, 
four PEMS datasets (PEMS03, PEMS04, PEMS07, PEMS08)~\citep{chen2001freeway},
Exchange, Weather, Traffic, Electricity (ECL)~\citep{wu2021autoformer}, and Solar-Energy (Solar)~\citep{lai2018modeling}.
Details of the dataset statistics are provided in 
Appendix~\ref{sec:data}.

\textbf{Baseline methods.}
We follow the baseline methods and results from S-Mamba \citep{wang2024mamba}.
For the baseline methods, we consider Transformer-based models, including iTransformer \citep{liu2023itransformer}, PatchTST \citep{nie2022time},  and Crossformer \citep{zhang2023crossformer},
as well as linear/MLP models, including TimesNet \citep{wu2022timesnet}, DLinear \citep{zeng2023transformers}, and RLinear \citep{li2023revisiting}. Additionally, we include
 S-Mamba \citep{wang2024mamba}, which is a Mamba-based TS forecasting model.
Details of the baseline methods are provided in Appendix~\ref{sec:baseline}.

\textbf{Experimental setups.}
We follow the experimental setups from iTransformer and S-Mamba.
Note that we do not tune any hyperparameters except for $\lambda$, which is related to the proposed regularization, while adhering to the values used in S-Mamba for all other hyperparameters concerning the model architecture and optimization.
For dataset splitting, we adhere to the standard protocol of dividing all datasets into training, validation, and test sets in chronological order.
Details of the experimental setups, including the size of the input window and the forecast horizon, are provided in Appendix~\ref{sec:data}.

\begin{table*}[t]
\vspace{-15pt}
\centering
\begin{adjustbox}{max width=1.023\textwidth}
\begin{NiceTabular}{c|cccc|cc|cc|cc|cc|cc|cc|cc}
\toprule
\multirow{4}{*}{Models} & \multicolumn{6}{c}{(1) Mamba} & \multicolumn{6}{c}{(2) Transformer} & \multicolumn{6}{c}{(3) Linear/MLP}   \\
\cmidrule(lr){2-7} \cmidrule(lr){8-13} \cmidrule(lr){14-19}
& \multicolumn{4}{c}{SOR-Mamba} & \multicolumn{2}{c}{\multirow{2.5}{*}{S-Mamba}} & \multicolumn{2}{c}{\multirow{2.5}{*}{iTransformer}} & \multicolumn{2}{c}{\multirow{2.5}{*}{PatchTST}} & \multicolumn{2}{c}{\multirow{2.5}{*}{Crossformer}} & \multicolumn{2}{c}{\multirow{2.5}{*}{TimesNet}} & \multicolumn{2}{c}{\multirow{2.5}{*}{DLinear}} & \multicolumn{2}{c}{\multirow{2.5}{*}{RLinear}}  \\
\cmidrule(lr){2-5}
&\multicolumn{2}{c}{FT}&\multicolumn{2}{c}{SL}& \\
\cmidrule(lr){1-1} \cmidrule(lr){2-3}\cmidrule(lr){4-5}\cmidrule(lr){6-7}\cmidrule(lr){8-9} \cmidrule(lr){10-11}\cmidrule(lr){12-13}\cmidrule(lr){14-15}\cmidrule(lr){16-17}\cmidrule(lr){18-19}
 Metric & MSE & MAE & MSE & MAE & MSE & MAE & MSE & MAE & MSE & MAE & MSE & MAE & MSE & MAE & MSE & MAE & MSE & MAE \\
\midrule

ETTh1 & \first{.433}&\second{.436}&\second{.442}& {.438}&.457 & .452 & .454 & {.449} & .469 & .454 & .529 & .522 &  .458 & .450 & .456 & .452 & {.446} & \first{.434} \\

ETTh2 & \second{.376}&\second{.405}&.382&.407&\fourth{.383} & \fourth{.408} & .384 & .407 & .387 & .407 & .942 & .684 &  .414 & .427 & .559 & .515 & \first{.374} & \first{.398} \\

ETTm1 & \second{.391}&\first{.400}&.396&\second{.401}&\fourth{.398} & \fourth{.407} & .408 & .412 & \first{.387} & \first{.400} & .513 & .496 &  .400 & .406 & .403 & .407 & .414 & .407  \\

ETTm2 & \first{.281}&\second{.327}&\second{.284}&.329&.290 & .333 & .293 & .337 & \first{.281} & \first{.326} & .757 & .610 &  .291 & .333 & .350 & .401 & \fourth{.286} & \second{.327}  \\

PEMS03 &\first{.121}&\first{.227}&.137&.242& \second{.133} & \second{.240} & {.142} & {.248} & .180 & .291 & .169 & .281 & .147 & .248 & .278 & .375  & .495 & .472 \\

PEMS04&\second{.099}&\first{.203}&.107&.212& \first{.096} & \second{.205} & \fourth{.121} & \fourth{.232}& .195 & .307 & .209 & .314 &  .129 & .241 & .295 & .388  & .526 & .491 \\

PEMS07 &\first{.088}&\first{.186}& .091&\second{.191}& \second{.090} & \second{.191} & \fourth{.102} & \fourth{.205}  & .211 & .303 & .235 & .315 &.124 & .225 & .329 & .395 & .504 & .478 \\

PEMS08 &\first{.142}&\first{.232}&\third{.162}&.247& \second{.157} & \second{.242} & {.254} & {.306}  & .280 & .321 & .268 & .307 & .193 & .271 & .379 & .416 & .529 & .487 \\

Exchange & \second{.358}&\first{.402}&.363&{.405}&.364 & .407 & \fourth{.368} & {.409} & .367 & \second{.404} & .940 & .707 &  .416 & .443 & \first{.354} & .414 & .378 & .417 \\

Weather & \second{.256}&\first{.277}&.257&\second{.278}&\first{.252} & \first{.277} & \fourth{.260} & {.281}  & .259 & .281 & .259 & .315 &  .259 & .287 & .265 & .317 & .272 & .291 \\

Solar & \first{.230}&\first{.259}&.242&.274&\fourth{.244} & \fourth{.275} & \second{.234} & \second{.261}& .270 & .307 & .641 & .639 &  .301 & .319 & .330 & .401  & .369 & .356  \\

ECL & \first{.168}&\second{.264}&\second{.169}&\first{.262}&\third{.174} & \third{.269} & \fourth{.179} & \fourth{.270}  & .205 & .290 & .244 & .334 &  .192 & .295 & .212 & .300 & .219 & .298 \\

Traffic &\first{.402}&\first{.273}&\second{.412}&\second{.276}& \third{.417} & \third{.277} & \fourth{.428} & \fourth{.282}  & .481 & .304 & .550 & .304 &  .620 & .336 & .625 & .383 & .626 & .378 \\

\midrule
\rowcolor{gray!20} Average
&\first{.257}&\first{.299}&\second{.265}&\second{.305}& \third{.266} & \third{.307} & \fourth{.278} & \fourth{.315}  & .306 & .338 & .481 & .448 & .303 & .329 & .372 & .397 & .418 & .403 \\
\midrule
\rowcolor{gray!20}1$^{\text{st}}$ Count & \first{33} & \first{31} & 7 & \second{10} & \second{10} & 7 & 1 & 3 & 8& 7& 3 & 0  & 0 & 0 & 2 & 0 & 3& 9 \\
\rowcolor{gray!20}2$^{\text{nd}}$ Count & \second{15} & \first{19} & \first{18} & \first{19} & 13 & \second{13} & 9 & 6 & 1& 6& 0 & 0  & 0 & 1 & 2 & 0 & 2& 2 \\
\bottomrule
\end{NiceTabular}
\end{adjustbox}
\caption{\textbf{Results of multivariate TS forecasting.}
We compare our method with the SOTA methods under both SL and SSL settings.
The best results are in \first{bold} and the second best are \second{underlined}.}
\label{tbl:tsf}
\vspace{-5pt}
\end{table*}

\subsection{Time Series Forecasting}
Table~\ref{tbl:tsf} 
presents the comprehensive results for the multivariate TS forecasting task, showing the average MSE/MAE across four horizons over five runs.
The results demonstrate that our proposed SOR-Mamba outperforms the SOTA Transformer-based models and S-Mamba, which uses the bidirectional Mamba, whereas our approach utilizes the unidirectional Mamba, providing greater efficiency as discussed in Table~\ref{tbl:eff}. 
Furthermore, self-supervised pretraining (SSL) with CCM yields additional performance gains compared to the supervised setting (SL),
with comparisons to SL and SSL (LP and FT) shown in Table~\ref{tbl:lpft}.
Full results of Table~\ref{tbl:tsf} are provided in 
Appendix~\ref{sec:full}.

\subsection{Transfer Learning}
To assess the transferability of our method, we conduct transfer learning experiments in both in-domain and cross-domain transfer settings following SimMTM~\citep{dong2023simmtm}, where source and target datasets share the same frequency in the in-domain setting, while they do not in the cross-domain setting.
Table~\ref{tbl:transfer} 
presents the average MSE across four horizons, demonstrating that SOR-Mamba consistently outperforms S-Mamba, achieving nearly a 5\% performance gain in FT.

\begin{figure*}[t]
    \centering
    \begin{minipage}{.38\textwidth}
        \centering
        \begin{adjustbox}{max width=1.00\textwidth}
            \begin{NiceTabular}{c|ccc}
            \toprule
              \multirow{2.5}{*}{Dataset} & \multirow{2.5}{*}{SL} & \multicolumn{2}{c}{SSL (CCM)}  \\
              \cmidrule(lr){3-4} 
              & & LP & FT \\
            \toprule
            ETTh1 & \second{.442} & .452 & \first{.433} \\
            ETTh2 & \second{.382} & \first{.376} & \first{.376} \\
            ETTm1 & \second{.396} & .399 & \first{.391} \\
            ETTm2 & .284 & \second{.283} & \first{.281} \\
            Exchange & .363 & \first{.349} & \second{.358} \\
            Solar & \second{.242} & \first{.230} & \first{.230} \\
            ECL & \second{.169} & \second{.169} & \first{.168} \\
            \bottomrule
            \end{NiceTabular}
        \end{adjustbox}
        \captionsetup{type=table}
        \caption{SL vs. SSL.}
        \label{tbl:lpft}
    \end{minipage}
    \hfill
    \begin{minipage}{.565\textwidth}
               \centering
        \begin{adjustbox}{max width=1.00\textwidth}
            \begin{NiceTabular}{c|cc|ccc|ccc}
            \toprule
            & \multirow{2.5}{*}{Source} & \multirow{2.5}{*}{Target}  & \multicolumn{3}{c}{\text{SOR-Mamba}} & \multicolumn{3}{c}{\text{S-Mamba}} \\
            \cmidrule(lr){4-6} \cmidrule(lr){7-9}
            &  &  & SL & LP  & FT & SL & LP & FT \\ 
            \midrule
            \multirow{3.5}{*}{\shortstack{In-\\domain}}&ETTh2&ETTh1 &  \second{.442} &   .452 &   \first{.433} & .457 & .450 & .464 \\
            &ETTm2&ETTm1&   \second{.396} &   .401 &  \first{.390}& .398 & .398 & {.400}   \\
            \cmidrule(lr){2-9} 
            &\multicolumn{2}{c}{\cellcolor{gray!20} \text{Average}} &  \cellcolor{gray!20} \second{.419} & \cellcolor{gray!20} .427 &\cellcolor{gray!20} \first{.411} & \cellcolor{gray!20}.428 & \cellcolor{gray!20}.425 & \cellcolor{gray!20}.432  \\
            \midrule
            \multirow{7.5}{*}{\shortstack{Cross-\\domain}}&ETTm2&ETTh1&   \second{.442} & .448 &   \first{.433} &.457 &.450 & .455   \\
            &ETTh2&ETTm1&   \second{.396}&    {.399} &   \first{.391} &.398 & .401 & .402  \\
            &ETTm1&ETTh1 & \second{.442} &  .449 &  \first{.434} & .457& .450 & .468  \\
            &ETTh1&ETTm1&  \second{.396}&   .404 &  \first{.391} & .398 & {.403} & .399  \\
            &Weather&ETTh1&   \first{.442} &   {.545} &   {.542} & \second{.457} & .546 & .552   \\
            &Weather&ETTm1&  \first{.396}&   {.457} &  {.458} & \second{.398}  & .460 & .501  \\
            \cmidrule(lr){2-9} 
            & \multicolumn{2}{c}{\cellcolor{gray!20} \text{Average}} & \cellcolor{gray!20} \first{.419} & \cellcolor{gray!20} .450 & \cellcolor{gray!20} .441 & \cellcolor{gray!20} \second{.428} & \cellcolor{gray!20} .452 & \cellcolor{gray!20} .463 \\
            \bottomrule
            \end{NiceTabular}
        \end{adjustbox}
        \captionsetup{type=table}
        \caption{Results of transfer learning.}
 \label{tbl:transfer}
    \end{minipage}
    \begin{minipage}{1.00\textwidth}
    \centering
    \vspace{5pt}
    \begin{adjustbox}{max width=1.00\textwidth}
    \begin{NiceTabular}{l|cccc|cc|cc}
    \toprule
      \multirow{4}{*}{Improvements} & \multicolumn{6}{c}{(1) Performance} & \multicolumn{2}{c}{(2) Efficiency}\\
      \cmidrule(lr){2-7} \cmidrule(lr){8-9}
       & \multicolumn{4}{c}{Average MSE across four horizons} & \multicolumn{2}{c}{Average} & \multirow{2.5}{*}{\# Params.} & \multirow{2.5}{*}{Impr.}\\
      \cmidrule(lr){2-5} \cmidrule(lr){6-7} 
       & ETTh1 & ETTh2 & ETTm1 & ETTm2 & MSE & Impr.& \\
      \midrule
      S-Mamba & .457 & .383 & .398 & .290 & .382 & - & 9.29M & - \\
      \rowcolor{green!20} + Regularization & .452 & \second{.382}  & \second{.394} & .286 & .378 & 1.0\% & 9.29M & - \\
      \rowcolor{blue!10}+ Bi $\rightarrow$ Unidirectional & .449 & \second{.382}  & .396 & .285 & .378 & 0.1\% & \second{5.81M} & 37.5\% \\
      \rowcolor{blue!10}+ Remove 1D-conv & \second{.442} & \second{.382} & {.396} & \second{.284} & \second{.376} & 0.5\% & \first{5.80M} & 0.1\% \\
      \rowcolor{yellow!20} + CCM & \first{.433} & \first{.376} & \first{.391} & \first{.281}  & \first{.370} & 1.5\% &  \first{5.80M} & -\\
      \bottomrule
    \end{NiceTabular}
    \end{adjustbox}
\captionsetup{type=table}
\caption{Ablation study of \textbf{\colorbox{green!20}{Regularization}}, \textbf{\colorbox{blue!10}{Model architecture}} and \textbf{\colorbox{yellow!20}{Pretraining task}}.}
\vspace{-10pt}
\label{tbl:ablation}
\end{minipage}
\end{figure*}

\subsection{Ablation Study}
To demonstrate the effectiveness of our method, 
we conduct an ablation study 
using four ETT datasets
to evaluate the impact of the following components: 
1) adding the regularization term, 
2) using the unidirectional Mamba instead of the bidirectional Mamba, 
3) removing the 1D-conv, 
and 4) pretraining with CCM.
Table~\ref{tbl:ablation} presents the results, indicating that using all proposed components results in the best performance 
and that our method outperforms S-Mamba
with 37.6\% fewer model parameters.
The full results of the ablation study are provided in Appendix~\ref{sec:ablation}.

\section{Analysis}
\textbf{Sequential order bias.}
The degree of a sequential order bias may vary depending on the characteristics of the datasets. We 
consider two factors affecting this degree: 
1) the \textit{correlation between channels} and 2) the \textit{number of channels} in the dataset. 
To evaluate the relationships between these factors and the degree of bias,
we quantify the degree of a sequential order bias for each dataset 
by measuring the difference in performance (average MSE across four horizons) when the channel order is reversed, 
using SOR-Mamba without regularization.
\setlength{\intextsep}{1.2pt} 
\setlength{\abovecaptionskip}{1.2pt}
\begin{wrapfigure}{r}{0.39\textwidth}
  \centering
  \includegraphics[width=0.39\textwidth]{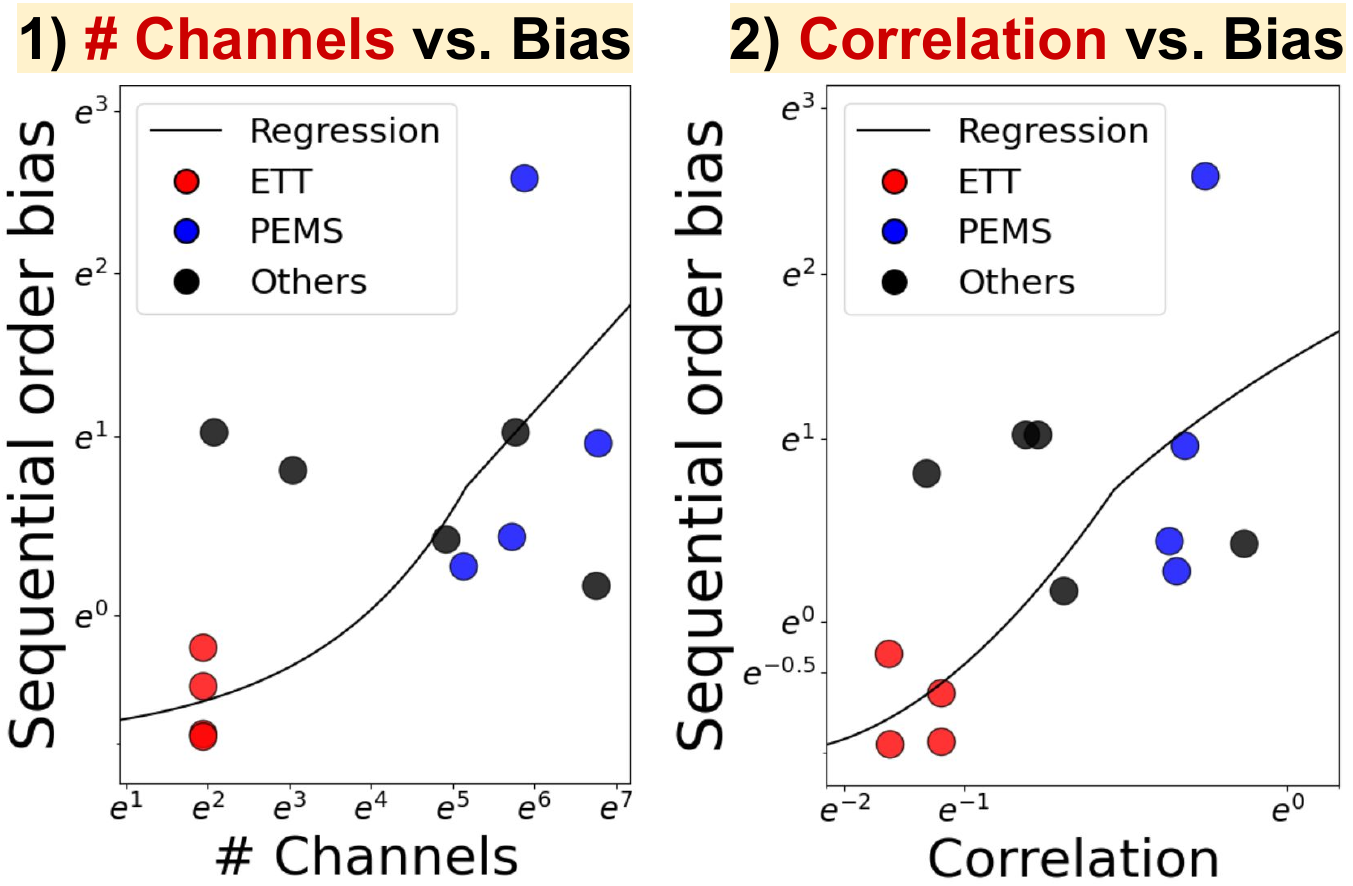}
    \caption{Varying bias across datasets.}    
  \label{fig:cob}
\end{wrapfigure}
Figure~\ref{fig:cob} shows the results with two plots, 
where the x-axes represent the number of channels 
and correlation between the channels
(i.e., average of the off-diagonal elements in the correlation matrix\footnote{We use its absolute value, as high correlation does not always indicate a strong relationship, 
with strong negative relationships near $-1$. Additionally, we use only the off-diagonal elements to exclude autocorrelation.} between the channels)%
,
and the y-axes represent the degree of a sequential order bias, 
with all axes shown on a log scale.
The results show that the bias increases 1) as the channels become more correlated and 2) as the number of channels increases.
For example, four ETT datasets
containing seven channels with low correlation show low bias, 
whereas four PEMS datasets
containing over 100 channels with high correlation exhibit high bias.

\setlength{\abovecaptionskip}{6pt}
\begin{figure*}[t]
    \vspace{-20pt}
    \centering
    \begin{minipage}{1.00\textwidth}
        \centering
        \begin{adjustbox}{max width=1.00\textwidth}
        \begin{NiceTabular}{cc|ccccccccccccc}
        \toprule
        \multicolumn{2}{c}{Mamba}
        & \multicolumn{4}{c}{ETT} & \multicolumn{4}{c}{PEMS} & \multirow{2.5}{*}{Exchange} & \multirow{2.5}{*}{Weather} & \multirow{2.5}{*}{Solar} & \multirow{2.5}{*}{ECL} & \multirow{2.5}{*}{Traffic} \\
        \cmidrule(lr){1-2} \cmidrule(lr){3-6}\cmidrule(lr){7-10}
         \# & Reg. & h1 & h2 & m1 & m2 & 03 & 04 & 07 & 08 & &&&&\\
        \midrule
        \multirow{2}{*}{Bi} & \xmark & .457 & .383 & .398 & .290 & .133 & \first{.096} & \first{.090} & .157 & .364 & \first{.252} & \first{.244} & .174 & .417 \\
          & \cellcolor{gray!20} \cmark & \cellcolor{gray!20} \first{.452} & \cellcolor{gray!20} \first{.382} & \cellcolor{gray!20} \first{.394} & \cellcolor{gray!20} \first{.286} & \cellcolor{gray!20} \first{.131} & \cellcolor{gray!20} \first{.096} & \cellcolor{gray!20} .092 & \cellcolor{gray!20} \first{.155} & \cellcolor{gray!20} \first{.361} & \cellcolor{gray!20} \first{.252} & \cellcolor{gray!20} {.245} & \cellcolor{gray!20} \first{.170} & \cellcolor{gray!20} \first{.411} \\
        \midrule
        \multirow{2}{*}{Uni} & \xmark & .455 & .383 & .403 & .289 & .140 & .102 & .094 & .161 & .364 & \first{.255} & \first{.244} & .175 & \first{.416}  \\
          & \cellcolor{gray!20} \cmark & \cellcolor{gray!20} \first{.449} & \cellcolor{gray!20} \first{.382} & \cellcolor{gray!20} \first{.396} & \cellcolor{gray!20} \first{.285} & \cellcolor{gray!20} \first{.135} & \cellcolor{gray!20} \first{.101} & \cellcolor{gray!20} \first{.091} & \cellcolor{gray!20} \first{.158} & \cellcolor{gray!20} \first{.361} & \cellcolor{gray!20} \first{.255} & \cellcolor{gray!20} \first{.244} & \cellcolor{gray!20} \first{.171} & \cellcolor{gray!20} \first{.416}  \\
        \bottomrule
        \end{NiceTabular}
        \end{adjustbox}
        \captionsetup{type=table}
        \caption{\textbf{Effect of regularization.} Regularization enhances both the unidirectional and the bidirectional Mamba. Note that we do not remove the 1D-conv to isolate the effect of regularization.}
        \label{tbl:reg}
    \end{minipage}
    \hfill
    \begin{minipage}{1.00\textwidth}
        \centering
        \vspace{10pt}
        \begin{adjustbox}{max width=1.00\textwidth}
        \begin{NiceTabular}{cc|ccccccccccccc}
        \toprule
        \multicolumn{2}{c}{Mamba}
        & \multicolumn{4}{c}{ETT} & \multicolumn{4}{c}{PEMS} & \multirow{2.5}{*}{Exchange} & \multirow{2.5}{*}{Weather} & \multirow{2.5}{*}{Solar} & \multirow{2.5}{*}{ECL} & \multirow{2.5}{*}{Traffic} \\
        \cmidrule(lr){1-2} \cmidrule(lr){3-6}\cmidrule(lr){7-10}
         \# & 1D-conv & h1 & h2 & m1 & m2 & 03 & 04 & 07 & 08 & &&&&\\
        \midrule
        \multirow{2}{*}{Bi} & \cmark & .457& \first{.383}& .398& .290 & \first{.133} & \first{.096}& .090& .157& \first{.364} & \first{.252}& .244& .174& .417 \\
         & \cellcolor{gray!20} \xmark & \cellcolor{gray!20} \first{.441}& \cellcolor{gray!20} \first{.383} & \cellcolor{gray!20} \first{.396} & \cellcolor{gray!20} \first{.285}& \cellcolor{gray!20} .137& \cellcolor{gray!20} .102 & \cellcolor{gray!20} \first{.089} & \cellcolor{gray!20} \first{.148} & \cellcolor{gray!20} \first{.364}& \cellcolor{gray!20} .255 & \cellcolor{gray!20} \first{.242} &\cellcolor{gray!20} \first{.167} & \cellcolor{gray!20} \first{.414}  \\
        \midrule
        \multirow{2}{*}{Uni} & \cmark & .449 & \first{.382}& \first{.396} & .285 & \first{.135} & \first{.101}& \first{.091} & \first{.158}& \first{.361} & \first{.255}& .244& .171& .416 \\
           & \cellcolor{gray!20} \xmark & \cellcolor{gray!20} \first{.442}& \cellcolor{gray!20} \first{.382} & \cellcolor{gray!20} \first{.396} & \cellcolor{gray!20} \first{.284}& \cellcolor{gray!20} .137& \cellcolor{gray!20} .107 & \cellcolor{gray!20} \first{.091} & \cellcolor{gray!20} .162 & \cellcolor{gray!20} {.363}& \cellcolor{gray!20} .257 & \cellcolor{gray!20} \first{.242} & \cellcolor{gray!20} \first{.169} & \cellcolor{gray!20} \first{.412}\\
        \bottomrule
        \end{NiceTabular}
        \end{adjustbox}
        \captionsetup{type=table}
        \caption{\textbf{Effect of 1D-conv.}
        Removing the 1D-conv, which captures the local information within adjacent channels, improves the performance on
        TS datasets that lack a sequential order in channels.}
        \label{tbl:1dremove}
    \end{minipage}
    \vspace{-20pt}
\end{figure*}

\textbf{Effect of regularization.}
To validate the effect of the regularization strategy, we apply it to both the unidirectional and the bidirectional Mamba without removing the 1D-conv to isolate the effect of regularization.
The results are shown in Table~\ref{tbl:reg}, which presents the TS forecasting results of the average MSE across four horizons. 
These results indicate that 
it
not only improves the performance of the unidirectional Mamba 
but also benefits the bidirectional Mamba, which handles the sequential order bias through bidirectional scanning, making regularization complementary to this approach.

\setlength{\intextsep}{0.8pt} 
\begin{wrapfigure}{r}{0.30\textwidth}
  \centering
  \includegraphics[width=0.30\textwidth]{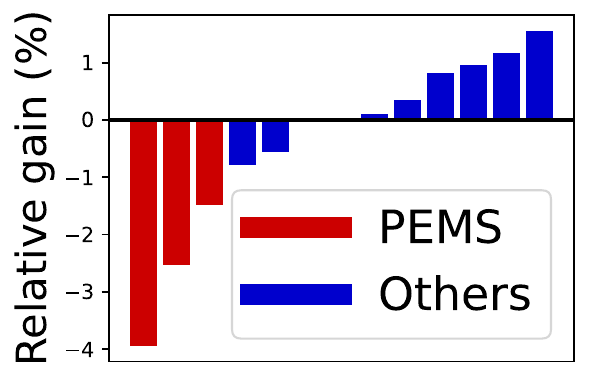}
    \caption{Effect of 1D-conv.}
  \label{fig:pems}
  \vspace{-4pt}
\end{wrapfigure}
\textbf{Effect of 1D-conv.} 
To demonstrate the unnecessity of the 1D-conv in Mamba for capturing CD, 
we remove it from both the unidirectional and the bidirectional Mamba, 
with the results of the average MSE across four horizons shown in Table~\ref{tbl:1dremove}. 
The results indicate that removing the 1D-conv, 
which captures the local information within nearby channels, 
improves the performance on general TS datasets where channels lack a sequential order. 
However, its removal may negatively impact datasets with ordered channels 
such as 
PEMS datasets~\citep{liu2022scinet},
which consist of traffic sensor data.
Figure~\ref{fig:pems} illustrates the relative gain from removing the 1D-conv in SOR-Mamba,
showing that three out of four PEMS datasets achieve better results with the 1D-conv than without it.

\begin{figure*}[t]
    \centering
    \vspace{-16pt}
    \begin{minipage}{.345\textwidth}
        \centering
        \begin{adjustbox}{max width=1.00\textwidth}
        \begin{NiceTabular}{c|ccc}
        \toprule
        \multirow{2.5}{*}{Dataset} &   \multirow{2.5}{*}{SL} & \multicolumn{2}{c}{SSL (CCM)}  \\
          \cmidrule(lr){3-4} 
          & & Global & Local \\
        \toprule
        ETTh1 & \second{.442} & {.445} & \first{.433} \\
        ETTh2 & .382 & \second{.380} & \first{.376} \\
        ETTm1 & .396 & \second{.393} & \first{.391} \\
        ETTm2 & .284 & \second{.283} & \first{.281} \\
        PEMS03 & .137 &  \second{.125} &  \first{.121} \\
        PEMS04 & .107 & \second{.101} & \first{.099} \\
        PEMS07 & \second{.091} & \first{.088} & \first{.088} \\
        PEMS08 & .162 & \second{.146} & \first{.142} \\
        Exchange & .363 & \second{.361} & \first{.358} \\
        Weather & \second{.257} & .258 & \first{.256} \\
        Solar & .242 & \first{.228} & \second{.230} \\
        ECL & \second{.169} & .170 & \first{.168} \\
        Traffic & .412 & \second{.410} & \first{.402} \\
        \midrule
        Average & .265 & \second{.260} & \first{.257}\\
        \bottomrule
        \end{NiceTabular}
        \end{adjustbox}
        \captionsetup{type=table}
        \caption{Global vs. Local corr.}
        \label{tbl:global_local}
    \end{minipage}
    \hfill
    \begin{minipage}{.62\textwidth}
        \centering
        \begin{adjustbox}{max width=1.00\textwidth}
        \begin{NiceTabular}{c|cccc|cccc}
        \toprule
        \multirow{4}{*}{Dataset} & 
        \multicolumn{4}{c}{SOR-Mamba} &
        \multicolumn{4}{c}{S-Mamba} 
        \\
        \cmidrule(lr){2-5} \cmidrule(lr){6-9}
          & \multirow{2.5}{*}{SL} & \multicolumn{3}{c}{SSL}  & \multirow{2.5}{*}{SL} & \multicolumn{3}{c}{SSL} \\
          \cmidrule(lr){3-5} \cmidrule(lr){7-9}
          & & Rec. & MM & CCM & & Rec. & MM &  CCM\\
        \toprule
        ETTh1 & .442 &\second{.434} & {.435} & \cellcolor{gray!20} \first{.433}  & \second{.457} & \first{.448} & \second{.457} & \cellcolor{gray!20} \second{.457} \\
        ETTh2 & .382 &\second{.378} & {.381} & \cellcolor{gray!20} \first{.376} & {.383} & \second{.381} &{.383} & \cellcolor{gray!20} \first{.380} \\
        ETTm1 & .396 & \first{.390} &{.396} & \cellcolor{gray!20} \second{.391} & .398 & .400 &\second{.397} & \cellcolor{gray!20} \first{.396}  \\
        ETTm2 & {.284} & \first{.279} & {.284} & \cellcolor{gray!20} \second{.281} & .290 & \first{.283} &{.288} & \cellcolor{gray!20}\second{.286}   \\
        PEMS03 & .137 & \second{.126} & \first{.121} &  \cellcolor{gray!20} \first{.121} & .133 & \second{.120} &.130 & \cellcolor{gray!20}\first{.119}  \\
        PEMS04 & .107 & .111 &\first{.095} & \cellcolor{gray!20} \second{.099}  & .096 & \first{.092} &.103 & \cellcolor{gray!20}\second{.093}  \\
        PEMS07 & .091 &.091 & \second{.090} & \cellcolor{gray!20} \first{.088} & .090 & \second{.086} &{.089} & \cellcolor{gray!20}\first{.085}   \\
        PEMS08  & .162 &\first{.139} & .144 & \cellcolor{gray!20} \second{.142} & {.157} & \first{.136} &{.157} & \cellcolor{gray!20}\second{.138}\\
        Exchange  & .363 & \second{.361} & \second{.361} & \cellcolor{gray!20} \first{.358} &  {.364} & \second{.363} & .378 &  \cellcolor{gray!20}\first{.361} \\
        Weather & \second{.257} & \first{.256} & \first{.256} & \cellcolor{gray!20} \first{.256} & .252 & \first{.249} &{.251} & \cellcolor{gray!20}\second{.250} \\
        Solar & .242 & \second{.231} & \second{.231} & \cellcolor{gray!20} \first{.230}  & .244 &  \first{.230} &{.239} &  \cellcolor{gray!20}\second{.233}  \\
        ECL & \second{.169} & .172 & \second{.169} & \cellcolor{gray!20}\first{.168}  & \second{.174} & .175 &\second{.174} & \cellcolor{gray!20}\first{.170} \\
        Traffic & .412 & \second{.410} & \second{.410} & \cellcolor{gray!20}\first{.402}  & .417 & .450 &\second{.415} & \cellcolor{gray!20}\first{.414}  \\
        \midrule
        Average & .265& .260&\second{.259}&\cellcolor{gray!20}\first{.257}  &.266&\second{.263}&.266& \cellcolor{gray!20}\first{.260}\\
        \bottomrule
        \end{NiceTabular}
        \end{adjustbox}
        \captionsetup{type=table}
        \caption{Comparison of various SSL pretraining tasks.}
        \label{tbl:pretraining_tasks}
    \end{minipage}
\end{figure*}

\textbf{Correlation for CCM.}
To assess the impact of using different correlations for CCM, 
we consider two candidates: \textit{local correlation}, which refers to the correlation between the channels of the input TS, 
and \textit{global correlation}, 
which refers to the correlation between the channels of the entire TS dataset.
Table~\ref{tbl:global_local} shows that using the local correlation yields better performance compared to the global correlation, 
although both approaches still outperform the supervised setting (SL).

\textbf{Effect of CCM.} To demonstrate the effect of CCM, we compare it with two other widely used pretraining tasks: masked modeling (MM)\citep{zerveas2021transformer} with a masking ratio of 50\%, and reconstruction (Rec.)\citep{lee2023learning}, along with the supervised setting.
Table~\ref{tbl:pretraining_tasks} presents the results using two backbones, S-Mamba and SOR-Mamba, showing that CCM consistently outperforms the other tasks across both backbones.

\setlength{\intextsep}{1.2pt} 
\begin{wrapfigure}{r}{0.315\textwidth}
  \centering
  \vspace{-4.3pt}
  \includegraphics[width=0.315\textwidth]{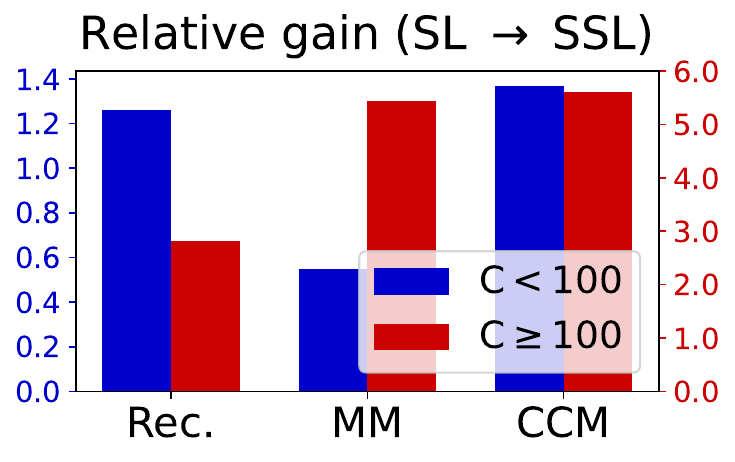}
  \vspace{-17.7pt}
    \caption{Comparison of SSL.}
  \label{fig:comparison_by_C}
\end{wrapfigure}
Furthermore, as CCM is designed to effectively capture CD in datasets, 
we compare the performance gain from three pretraining tasks based on the number of channels,
with six datasets containing fewer than 100 channels and seven datasets containing 100 or more channels.
Figure~\ref{fig:comparison_by_C} 
    shows the average performance gain from fine-tuning with the three tasks compared to SL, indicating that reconstruction is advantageous with fewer channels
and masked modeling excels with more channels, while CCM consistently outperforms in both cases.

\begin{figure*}[t]
    \centering
    \begin{minipage}{.415\textwidth}
        \centering
        \begin{adjustbox}{max width=1.00\textwidth}
        \begin{NiceTabular}{c|cc}
        \toprule
        $H$& SOR-Mamba  & S-Mamba \\
        \midrule
        96& $.378_{\pm \textbf{\textcolor{red}{.0003}}}$  & $.386_{\pm \textbf{.0010}}$  \\
        192 & $.428_{\pm \textbf{\textcolor{red}{.0002}}}$ 
 & $.440_{\pm \textbf{.0033}}$  \\
        336 & $.464_{\pm \textbf{\textcolor{red}{.0002}}}$ 
 & $.484_{\pm \textbf{.0046}}$  \\
        720 & $.464_{\pm \textbf{\textcolor{red}{.0004}}}$ 
 & $.502_{\pm \textbf{.0057}}$  \\
        \bottomrule
        \end{NiceTabular}
        \end{adjustbox}
        \captionsetup{type=table}
        \caption{Robustness to channel order.}
        \label{tbl:robust1}
    \end{minipage}
    \hfill
    \begin{minipage}{.57\textwidth}
        \vspace{-6pt}
        \centering
\includegraphics[width=1.000\textwidth]{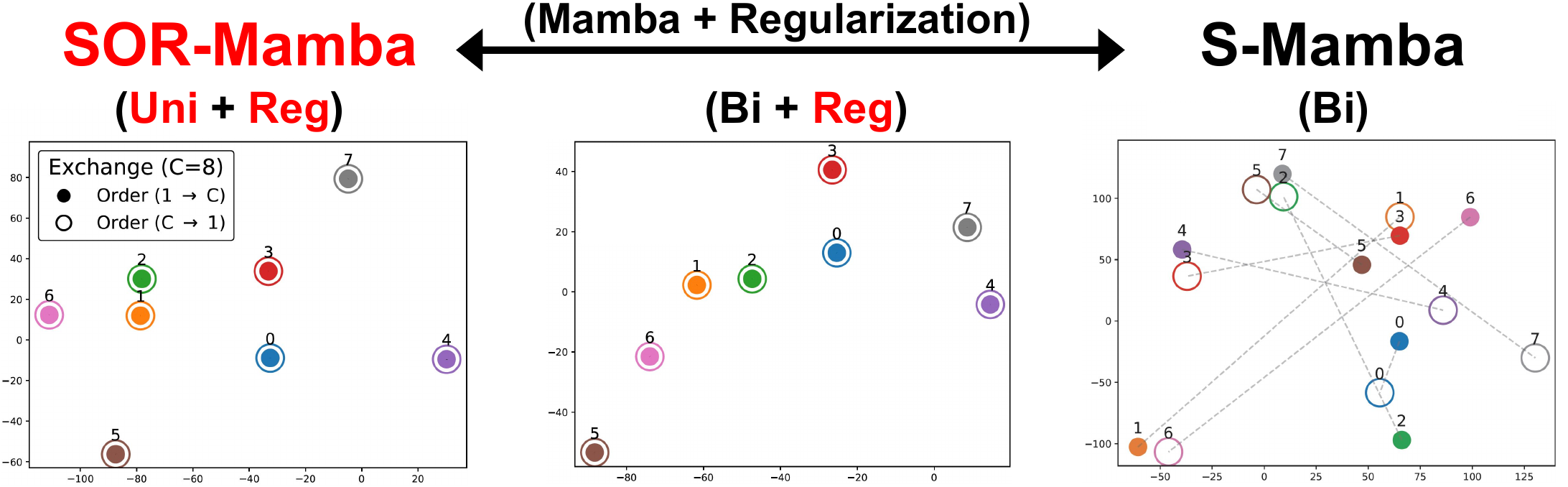}
        \caption{t-SNE of channel representations.}
        \label{fig:robust2}
    \end{minipage}
    \begin{minipage}{1.00\textwidth}
        \centering
        \vspace{8pt}
        \begin{adjustbox}{max width=1.00\textwidth}
        \begin{NiceTabular}{c|ccccccccccccc|c}
        \toprule
        \multirow{2.5}{*}{\shortstack{Architecture\\for TD}}
        & \multicolumn{4}{c}{ETT} & \multicolumn{4}{c}{PEMS} & \multirow{2.5}{*}{Exchange} & \multirow{2.5}{*}{Weather} & \multirow{2.5}{*}{Solar} & \multirow{2.5}{*}{ECL} & \multirow{2.5}{*}{Traffic} & \multirow{2.5}{*}{Avg.}\\
        \cmidrule(lr){2-5} \cmidrule(lr){6-9}
          & h1 & h2 & m1 & m2 & 03 & 04 & 07 & 08 & &&&&&\\
          \midrule
         - & \second{.446} & \second{.386} & \second{.397} & .286 & \second{.139} & \second{.109} & \second{.096} & \second{.164} & \first{.363} & \second{.258} & \second{.244} & \second{.170} & \second{.433} & \second{.268} \\
         Mamba & .447 & \second{.386} & .398 & \second{.285} & .140 & \second{.109}  & .097 & .165 & \first{.363} & .259 & .245 & .171 & .437 & .269 \\
        MLP & \first{.442}& \first{.382} & \first{.396} & \first{.284}&  \first{.137}& \first{.107} & \first{.091}& \first{.162} & \first{.363} & \first{.257} & \first{.242} & \first{.169} & \first{.412} & \first{.265}\\
        \bottomrule
        \end{NiceTabular}
        \end{adjustbox}
        \captionsetup{type=table}
        \caption{Various architectures for capturing TD.}
        \label{tbl:tdencoder}
    \end{minipage}
    \vspace{-20pt}
\end{figure*}

\setlength{\intextsep}{1.2pt} 
\setlength{\abovecaptionskip}{1.2pt}
\begin{figure*}[t]
    \vspace{-22pt}
    \centering
    \begin{minipage}{0.645\textwidth}
    \centering
    \begin{subfigure}{0.53\textwidth} 
    \includegraphics[width=1.000\textwidth]{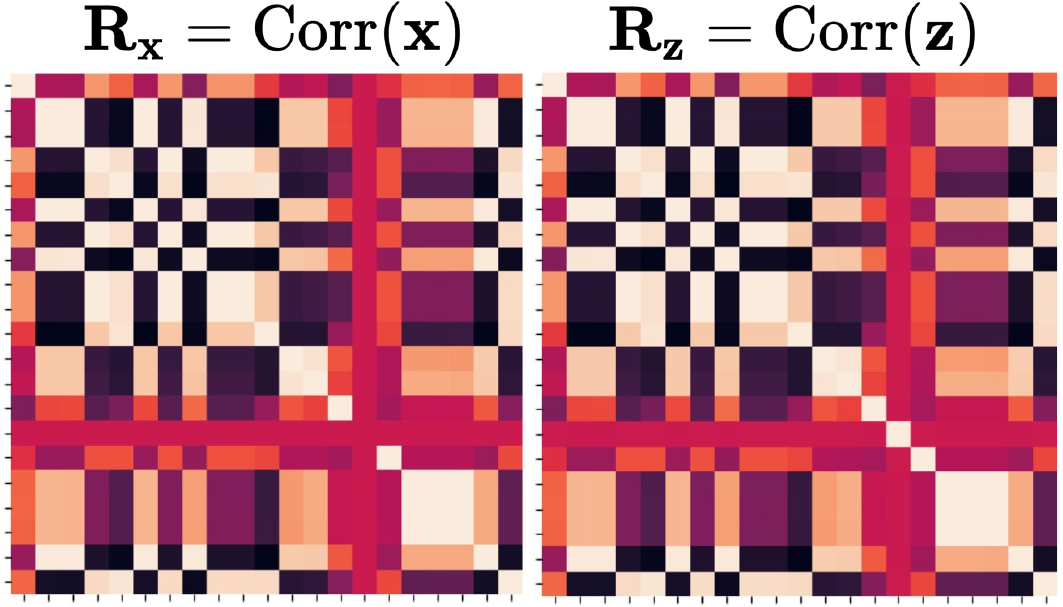}
    \caption{Visualization of $\mathbf{R_x}$ and $\mathbf{R_z}$.}
    \label{fig:preserve1}
    \end{subfigure}
    \begin{subfigure}{0.45\textwidth} 
    \includegraphics[width=1.000\textwidth]{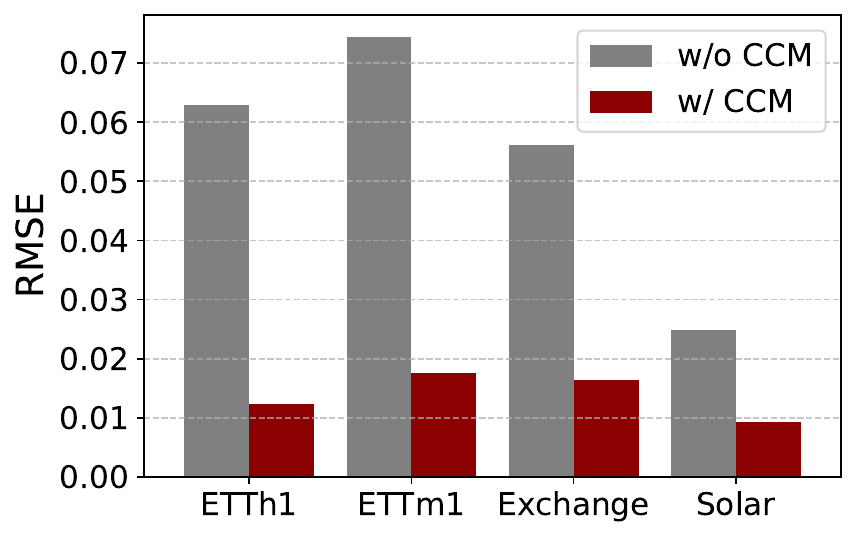}
    \caption{Comparison of $d(\mathbf{R_x},\mathbf{R_z})$.}
    \label{fig:preserve2}
    \end{subfigure}
    \vspace{2pt}
    \caption{Correlation matrices in data space and latent space.}
    \end{minipage}
    \begin{minipage}{0.325\textwidth}
        \centering
        \includegraphics[width=1.000\textwidth]{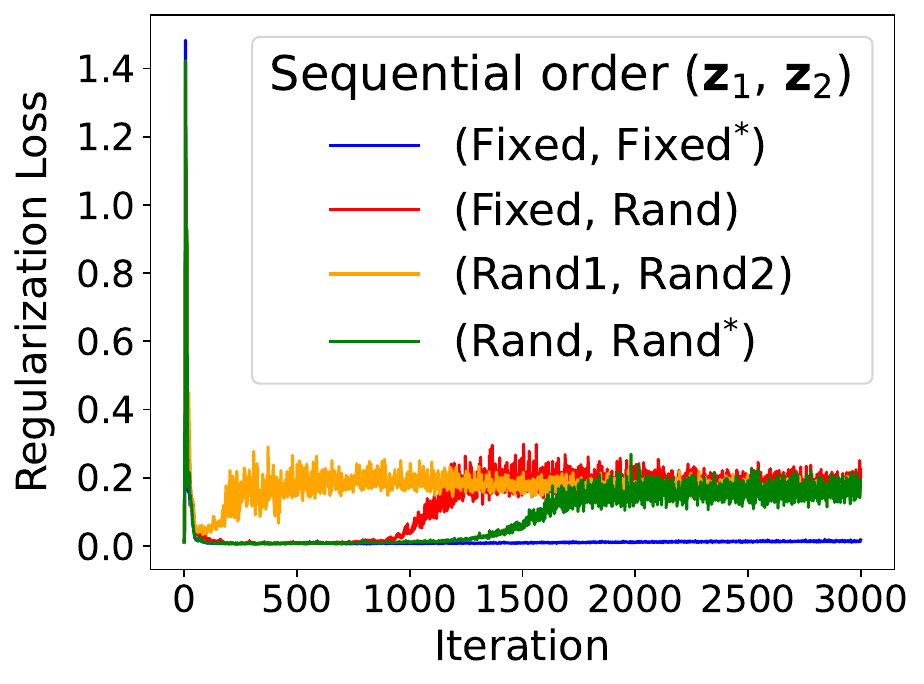}
        \caption{Regularization loss.}
        \label{fig:pems8_training}
    \end{minipage}
\end{figure*}
\textbf{Robustness to channel order.}
To demonstrate that 
our method
effectively addresses a sequential order bias, 
we conduct two analyses to show its robustness to the channel order.
First, we evaluate the performance variations 
with five random permutations of channel order using ETTh1,
where our method achieves a smaller standard deviation 
compared to S-Mamba,
as shown in Table~\ref{tbl:robust1}.
Additional results with different datasets are described in Appendix~\ref{sec:channel_order_robust}.
Second, we visualize the 
output tokens of the encoder
(i.e., embedding vectors of each channel)
using t-SNE~\citep{van2008visualizing} with Exchange.
Figure~\ref{fig:robust2} illustrates the results, 
showing that the tokens from the two views with reversed orders are consistent with regularization, while remaining inconsistent without it.

\textbf{Various architectures for TD.}
Following the recent studies~\citep{liu2023itransformer,wang2024mamba} that suggest employing simple models, e.g., MLPs, to capture TD in TS, we utilize an MLP 
for this purpose.
To examine the impact of different design choices of architecture
for capturing TD, 
we consider two alternatives: 
1) without employing any encoder for TD, 
and 2) using Mamba, 
following the 
previous work~\citep{wang2024mamba}. 
Table~\ref{tbl:tdencoder} shows the results, demonstrating that our method is robust to the choice of encoder for TD, 
achieving the best performance with an MLP.

\textbf{Correlation in the data space and the latent space.}
To demonstrate that CCM effectively preserves the relationships between channels from the data space to the latent space, we visualize the correlation matrices in both spaces 
with SOR-Mamba pretrained with CCM.
Figure~\ref{fig:preserve1} shows the results on the Weather dataset, which indicate that the relationships are effectively preserved with CCM. 
Additionally, we compare the distances between the matrices in both spaces, 
comparing SOR-Mamba without pretraining 
to the one pretrained with CCM. 
The results, illustrated in Figure~\ref{fig:preserve2}, show that the model pretrained with CCM exhibits a smaller difference between the matrices.

\textbf{Fixed vs. random order.}
To generate two embedding vectors 
for regularization, we explore 
four candidates based on whether 
the channel order of $\mathbf{z}_1$ and $\mathbf{z}_2$ are fixed or randomly permuted in each iteration.
Table~\ref{tbl:fixedvsrandom} shows the results with the average MSE across four horizons,
indicating that fixing the order yields better performance than permuting the order,
especially with a large number of channels ($C\geq 100)$.
We argue that fixing the order leads to stable training, while permuting the order results in instability, as shown in the regularization loss curves for PEMS08 in Figure~\ref{fig:pems8_training}.
Further analysis
regarding the channel order 
is discussed in Appendix~\ref{sec:fixed_vs_random}.

\textbf{Efficiency analysis.}
To demonstrate the efficiency of SOR-Mamba, we compare it with iTransformer and S-Mamba in terms of 1) the number of parameters, 2) memory usage, and 3) computational time. 
Table~\ref{tbl:eff} shows the results, indicating that SOR-Mamba outperforms these methods in all three aspects, particularly reducing the number of parameters by up to 38.1\% compared to S-Mamba. Note that the training time is measured per epoch, while the inference time is measured per data instance.

\begin{figure*}[t]
    \centering
    \begin{minipage}{.42\textwidth}
        \centering
        \begin{adjustbox}{max width=1.00\textwidth}
        \begin{NiceTabular}{c|cc|cccc|c}
        \toprule
        & \multicolumn{6}{c}{\colorbox{yellow!20}{$F$: Fixed}, \colorbox{blue!10}{$R$: Random}, $X^{\star}$: Reverse of $X$} & \multirow{4}{*}{\shortstack{Impr.\\(Robust.)}}\\
        \cmidrule{2-7}
        & \multirow{2}{*}{Order} & $\mathbf{z}_1$ & \colorbox{yellow!20}{$F$} & \colorbox{yellow!20}{$F$} & \colorbox{blue!10}{$R_1$} & \colorbox{blue!10}{$R$}  \\
        & &$\mathbf{z}_2$ & \colorbox{yellow!20}{$F^{\star}$} & \colorbox{blue!10}{$R$} & \colorbox{blue!10}{$R_2$} & \colorbox{blue!10}{$R^{\star}$} \\
        \midrule
        & Dataset & $C$ & (a) & (b) & (c) & (d) & (d) $\rightarrow$ (a) \\
        \cmidrule(lr){2-2} \cmidrule(lr){3-3} \cmidrule(lr){4-4} \cmidrule(lr){5-5} \cmidrule(lr){6-6} \cmidrule(lr){7-7} \cmidrule(lr){8-8}
        \multirow{6.5}{*}{\rotatebox{90}{$C<100$}} & ETTh1 & 7 &\first{.442} & \second{.443} & .446  & \second{.443} & 0.2\% \\ 
        & ETTh2 & 7 &\first{.382} & \first{.382} & \first{.382}  & \first{.382} & 0.0\% \\ 
        & ETTm1 & 7 &\first{.396} & \first{.396} & \first{.396}  & \first{.396} & 0.0\% \\ 
        & ETTm2 & 7 & \first{.284} & \second{.285} & \second{.285} & \second{.285} & 0.4\% \\  
        & Exchange & 8 &\first{.363} & \second{.364} & .365 & \second{.364} & 0.3\% \\ 
        & Weather & 21 &\first{.257} & \second{.258} & .260 & .260 & 1.2\% \\ 
        \cmidrule{2-8}
        &  \multicolumn{2}{c}{\cellcolor{gray!20} Average} &  \cellcolor{gray!20} \first{.354} & \cellcolor{gray!20} \second{.355} & \cellcolor{gray!20} .356 & \cellcolor{gray!20} \second{.355} & \cellcolor{gray!20} 0.3\% \\ 
        \midrule
        \multirow{7.5}{*}{\rotatebox{90}{$C \geq 100$}} & Solar & 137 &\first{.242} & \second{.245} & \second{.245} & .246 & 1.6\% \\ 
        & PEMS03 & 358 & \first{.137} & \second{.144} & .150 & .151 & 9.3\% \\ 
        & PEMS04 & 307 & \first{.107} & \second{.112} & .116 & .117 & 8.5\% \\ 
        & PEMS07 & 883 & \first{.091} & \second{.096} & .097 & \second{.096} & 5.2\% \\ 
        & PEMS08 & 170 & \first{.162} & \second{.163} & .169 & .172 & 5.8\% \\ 
        & ECL & 321 &\first{.169} & \second{.174} & .181 & .183 & 7.7\% \\ 
        & Traffic & 862 &\first{.412} & \second{.422} & .423 & .423 & 2.6\% \\ 
        \cmidrule{2-8}
        & \multicolumn{2}{c}{\cellcolor{gray!20} Average} & \cellcolor{gray!20} \first{.189} & \cellcolor{gray!20} \second{.194} & \cellcolor{gray!20} .197 & \cellcolor{gray!20} .198 & \cellcolor{gray!20} 4.9\% \\ 
        \bottomrule
        \end{NiceTabular}
        \end{adjustbox}
        \captionsetup{type=table}
        \caption{Channel orders for two views.}
        \label{tbl:fixedvsrandom}
    \end{minipage}
    \hfill
    \begin{minipage}{.558\textwidth}
        \centering
        \begin{adjustbox}{max width=1.00\textwidth}
        \begin{NiceTabular}{l|ccc|c}
        \toprule
        \multicolumn{1}{c}{Dataset: Traffic}& (a) & (b) & (c) & (b) $\rightarrow$ (c) \\ 
        \multicolumn{1}{c}{($L=96,H=96$)} & iTrans. & S-Mamba & SOR-Mamba &  Impr.\\
        \cmidrule{1-1} \cmidrule(lr){2-2} \cmidrule(lr){3-3} \cmidrule(lr){4-4} \cmidrule(lr){5-5} 
        \textbf{\# Parameters}\\
        \cmidrule(lr){1-1}
        In projector&0.05M & 0.05M & 0.05M& - \\
        Encoder-CD & 4.20M & 6.97M&\first{3.48M}& \first{50.1\%} \\
        Encoder-TD & 2.11M & 2.11M & 2.11M & - \\
        Out projector &0.05M & 0.05M & 0.05M & - \\
        \cmidrule(lr){2-4} \cmidrule(lr){5-5}
        Total &6.52M&9.29M&\first{5.80M}&\first{38.1\%}\\
        \midrule
        \textbf{Memory}\\
        \cmidrule(lr){1-1}
        Complexity &$\mathcal{O}\left(C^2\right)$& $\mathcal{O}\left(C\right)$& $\mathcal{O}\left(C\right)$ & - \\
        GPU memory (GB) &1.36&0.33&\first{0.32}&\first{4.2\%}\\
        \midrule
        \textbf{Computational time}\\
        \cmidrule(lr){1-1}
        Train (sec.)& 115.5 & 108.3 & \first{102.1} & \first{5.7\%}\\
        Inference (ms)& 14.6 & 9.9 & \first{8.7} & \first{11.3\%}\\
        \midrule
        Avg. MSE (four $H$) & 0.428 & 0.417 & \first{0.402} & \first{3.6\%} \\
        \bottomrule
        \end{NiceTabular}
        \end{adjustbox}
        \captionsetup{type=table}
        \caption{Efficiency analysis.}
        \label{tbl:eff}
    \end{minipage}
    \vspace{-15pt}
\end{figure*}

\setlength{\intextsep}{1.2pt} 
\setlength{\abovecaptionskip}{6pt}
\begin{wrapfigure}{r}{0.27\textwidth}
  \centering
  \includegraphics[width=0.27\textwidth]{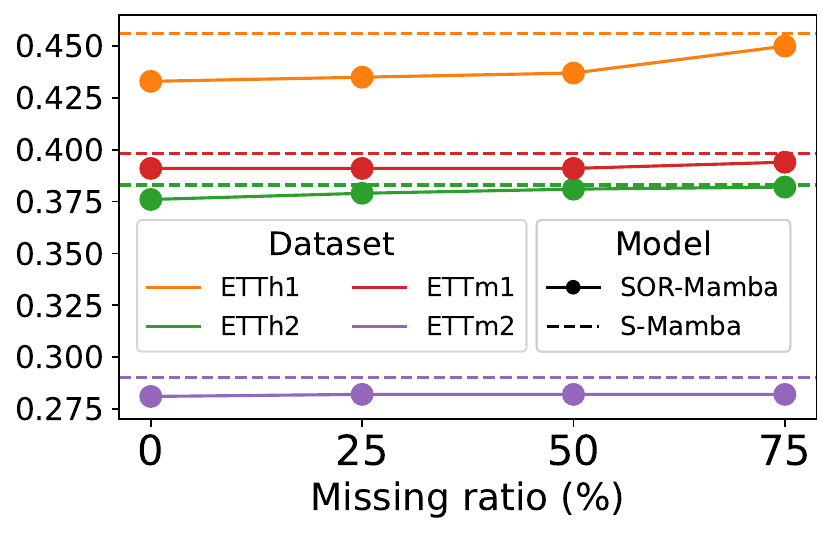}
    \vspace{-15pt}
    \caption{Missingness.}
    \vspace{-10pt}
    \label{fig:missing}
\end{wrapfigure}
\textbf{Robustness to missingness.}
To assess the robustness of our method in the presence of missing TS values,
we conduct experiments in scenarios where 25\%, 50\%, and 75\% of the TS values are randomly missing and interpolated using adjacent values.
Figure~\ref{fig:missing} shows the average MSE across four horizons, 
indicating that our method remains robust even with significant amounts of missing data 
and that
our method trained with missing values outperforms S-Mamba trained without missingness.

\section{Conclusion}
In this work, we introduce SOR-Mamba, a TS forecasting method that addresses the sequential order bias
by incorporating a regularization strategy and removing the 1D-conv from Mamba.
Additionally, we propose a novel pretraining task, CCM, to improve the model's ability to capture CD.
Our results demonstrate that the proposed method 
is robust to variations in channel order, leading to superior performance and greater efficiency in both standard and transfer learning scenarios. 
We hope that our work motivates further research on sequential order-robust Mamba in domains where a sequential order is not inherent, such as in tabular data.
\newpage

\bibliography{iclr2025_conference}
\bibliographystyle{iclr2025_conference}

\newpage
\appendix
\appendix
\numberwithin{table}{section}
\numberwithin{figure}{section}
\numberwithin{equation}{section}

\section{Dataset Statistics and Experimental Setups}
\label{sec:data}
\textbf{Dataset statistics.}
We assess the performance of SOR-Mamba across 13 datasets, with the dataset statistics detailed in Table~\ref{tab:dataset_stat},
where $C$ and $T$ denote the number of channels and timesteps, respectively.

\textbf{Experimental setups.}
We follow the same data processing steps and train-validation-test split protocol as used in S-Mamba~\citep{wang2024mamba}, maintaining a chronological order in the separation of training, validation, and test sets, using a 6:2:2 ratio for the Solar-Energy, ETT, and PEMS datasets, and a 7:1:2 ratio for the other datasets.
The results are shown in Table~\ref{tab:dataset_stat},
where $N$,$L$, and $H$ represent 
the dataset size, 
the size of the lookback window, 
and the size of the forecast horizon, respectively.
For all datasets and all models, $L$ is uniformly set to 96.
We do not tune any hyperparameters and adhere to those used in S-Mamba, except for $\lambda$, which is related to the proposed regularization, and is tuned using a grid search over $[0.001,0.01,0.1]$.

\begin{table*}[h]
\centering
\vspace{10pt}
\begin{adjustbox}{max width=1.00\textwidth}
\begin{NiceTabular}{l|c|c|c|c|c}
\toprule
\multirow{2.5}{*}{Dataset} & \multicolumn{2}{c}{Statistics} & \multicolumn{3}{c}{Experimental Setups} \\
\cmidrule(lr){2-3} \cmidrule(lr){4-6}
 & $C$ & $T$ & $(N_\text{train},N_\text{val},N_\text{test})$ & $L$ 
& $H$\\
\midrule
ETTh1 \citep{zhou2021informer} & \multirow{4}{*}{7} & 17420 & (8545, 2881, 2881) & \multirow{13}{*}{96} & \multirow{9}{*}{{\{96, 192, 336, 720\}}} \\
ETTh2 \citep{zhou2021informer} &  & 17420 & (8545, 2881, 2881)  &&\\
 ETTm1 \citep{zhou2021informer} &  & 69680 & (34465, 11521, 11521) && \\
 ETTm2 \citep{zhou2021informer} &  & 69680 & (34465, 11521, 11521) && \\
 \cmidrule{1-4}
Exchange \citep{wu2021autoformer} & 8 & 7588 & (5120, 665, 1422) &&\\
Weather \citep{wu2021autoformer} & 21 & 52696 & (36792, 5271, 10540)  &&\\
ECL \citep{wu2021autoformer}& 321 & 26304 & (18317, 2633, 5261)  &&\\
Traffic \citep{wu2021autoformer} & 862 & 17544 & (12185, 1757, 3509)  &&\\
Solar-Energy \citep{lai2018modeling} & 137 & 52560 & (36601, 5161, 10417)& & \\
\cmidrule{1-4}\cmidrule{6-6}
PEMS03 \citep{liu2022scinet} & 358 & 26209 & (15617, 5135, 5135) & & \multirow{4}{*}{{\{12, 24, 48, 96\}}}\\
PEMS04 \citep{liu2022scinet}& 307  & 15992 & (10172, 3375, 3375) && \\
PEMS07 \citep{liu2022scinet}& 883& 28224 & (16911, 5622, 5622) & & \\
PEMS08 \citep{liu2022scinet} & 170  & 17856 & (10690, 3548, 3548) & & \\
\bottomrule
\end{NiceTabular}
\end{adjustbox}
\caption{Datasets for TS forecasting.}
\label{tab:dataset_stat}
\end{table*}

\vspace{15pt}

\section{Baseline Methods} 
\label{sec:baseline}
\begin{itemize}[leftmargin=22pt, itemsep=4pt]
    \item S-Mamba~\citep{wang2024mamba}: S-Mamba utilizes the bidirectional Mamba to capture channel dependencies in TS by scanning the channels from both directions.
    \item PatchTST~\citep{nie2022time}: PatchTST segments TS into patches and feeds them into a Transformer in a channel independent manner.
    \item iTransformer~\citep{liu2023itransformer}: iTransformer reverses the conventional role of the Transformer in the TS domain by treating each channel rather than patches as a token, thereby emphasizing channel dependencies over temporal dependencies.
    \item Crossformer~\citep{zhang2023crossformer}: Crossformer employs a cross-attention mechanism to capture both temporal and channel dependencies in TS. 
    \item TimesNet~\citep{wu2022timesnet}: TimesNet captures both intraperiod and interperiod variations in 2D space using a parameter-efficient inception block.
    \item RLinear~\citep{li2023revisiting}: RLinear is a simple linear model that integrates reversible normalization and channel independence.
    \item DLinear~\citep{zeng2023transformers}: DLinear is a simple linear model with channel independent architecture, that employs TS decomposition.
\end{itemize}

\newpage
\section{S-Mamba vs. SOR-Mamba}
\label{sec:smamba_vs_SOR-Mamba}

\begin{figure*}[h]
\vspace{15pt}
\centering
\includegraphics[width=.95\textwidth]{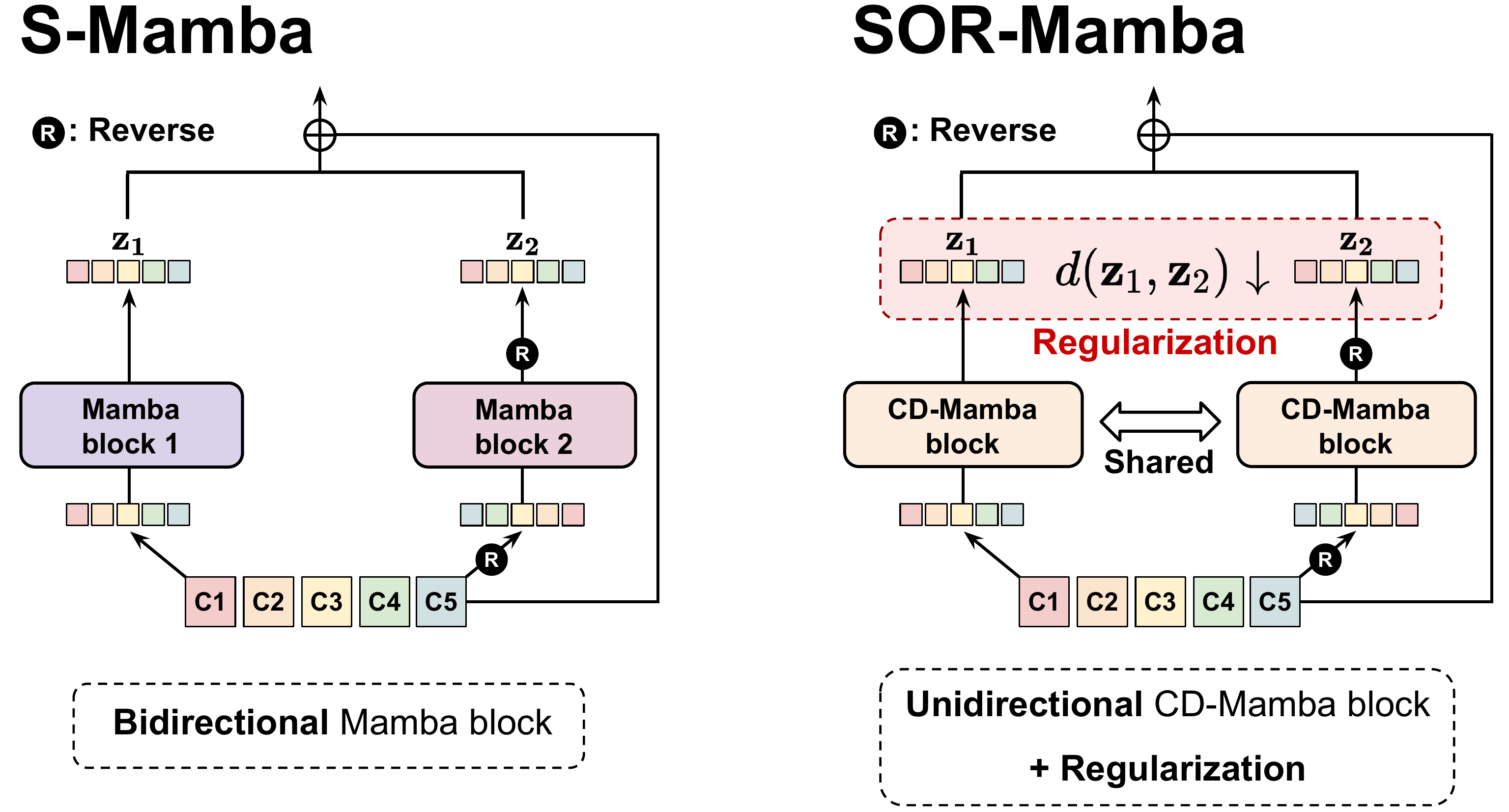} 
\caption{Comparison of S-Mamba and SOR-Mamba.}
\label{fig:smamba_vs_SOR-Mamba}
\end{figure*}

Figure~\ref{fig:smamba_vs_SOR-Mamba} visualizes the comparison between S-Mamba~\citep{wang2024mamba}, which employs the bidirectional Mamba to capture CD, and our method, SOR-Mamba, which uses a single unidirectional Mamba with regularization to capture CD.

\newpage
\section{Removal of 1D-Convolution}
\label{sec:1d_conv_remove}
\begin{figure*}[h]
\vspace{15pt}
\centering
\includegraphics[width=.83\textwidth]{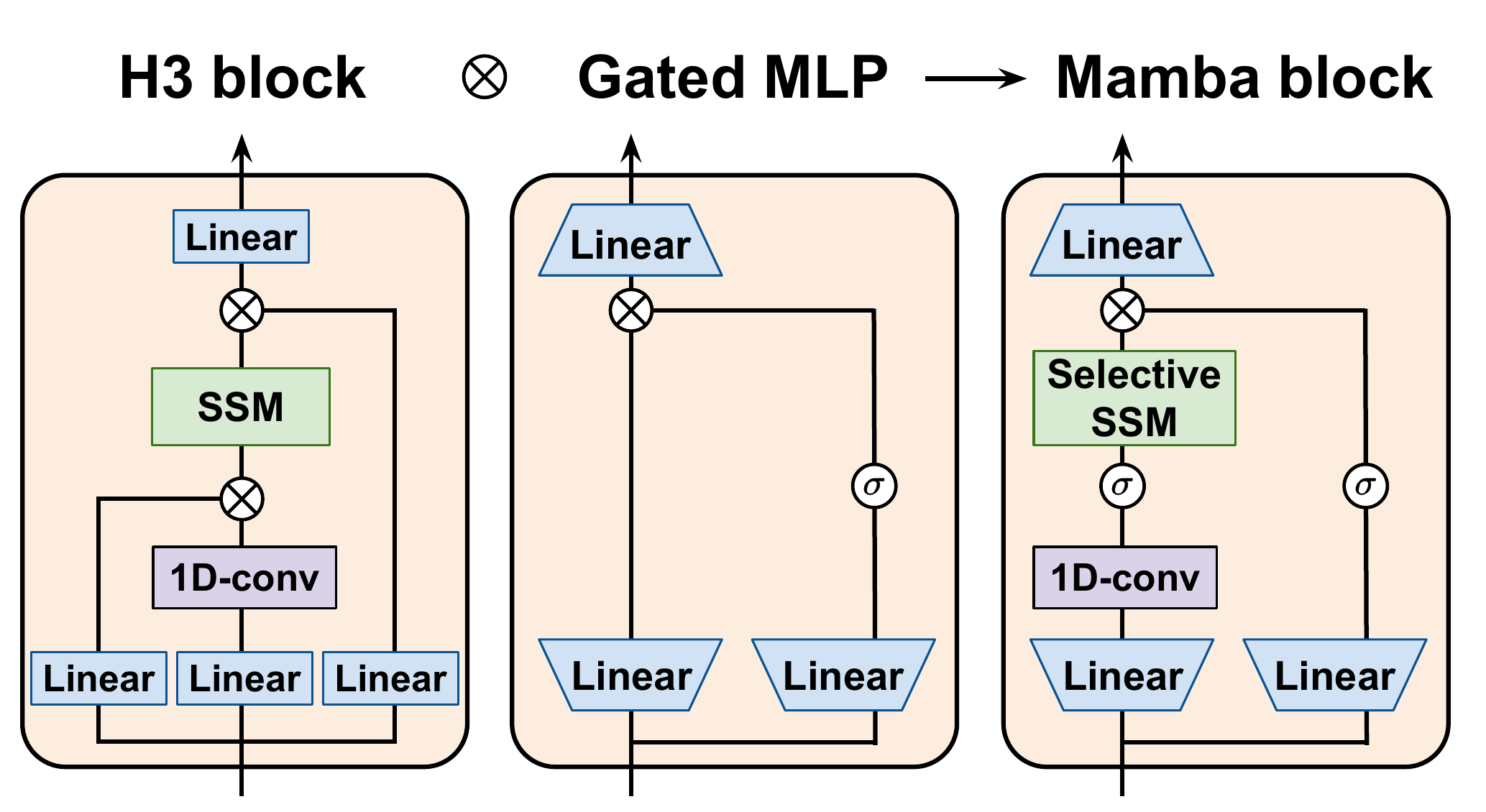} 
\caption{\textbf{Architecture of the original Mamba block.} 
The original Mamba block
contains 1D-conv before the SSM layer to capture local information within nearby tokens.1
}
\label{fig:cdmamba2}
\end{figure*}

\begin{figure*}[h]
\vspace{15pt}
\centering
\includegraphics[width=.83\textwidth]{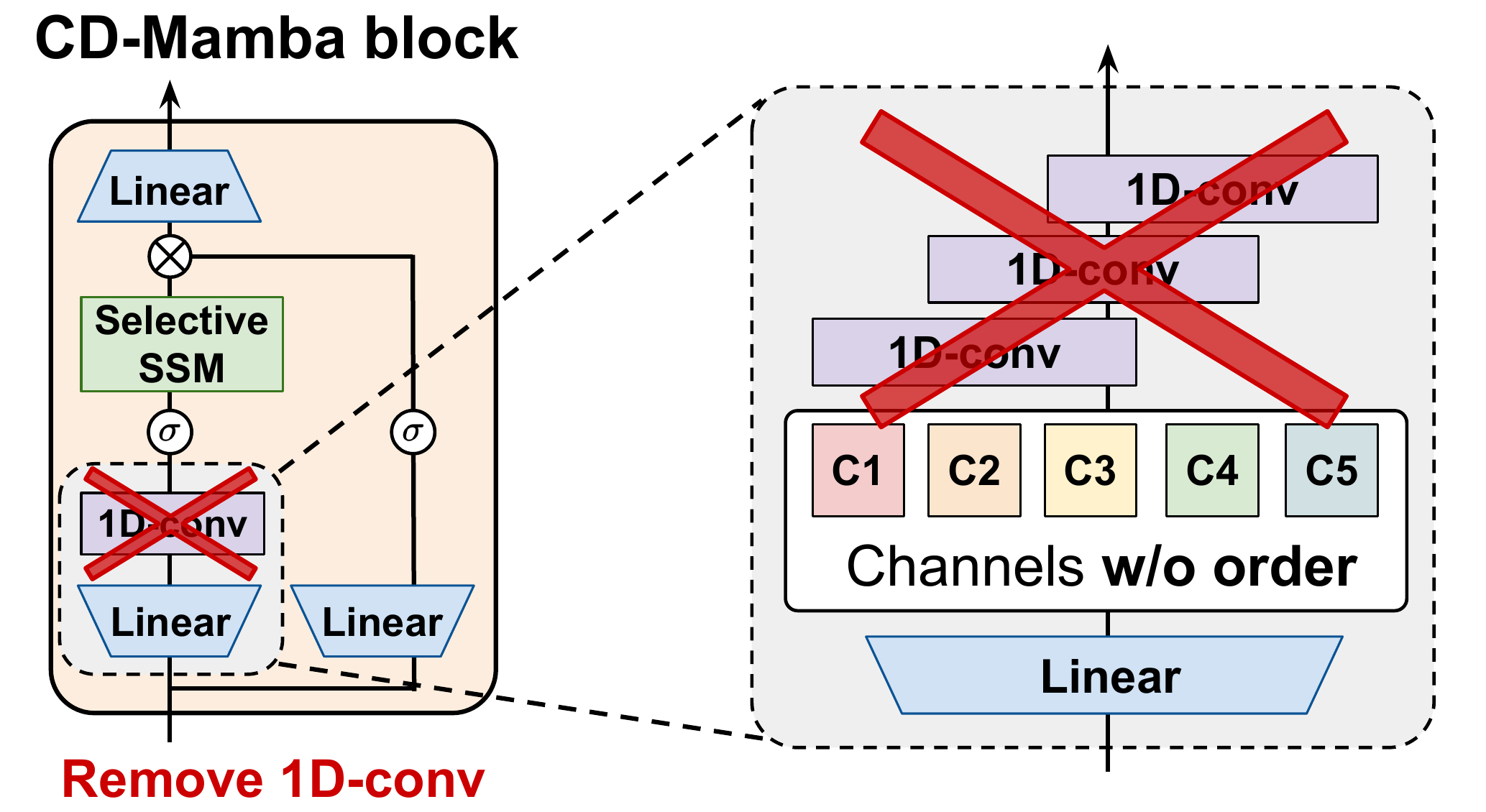} 
\caption{\textbf{Architecture of the CD-Mamba block.} 1D-conv before the selective SSM is removed from the original Mamba block, as the channels do not have a sequential order.}
\label{fig:cdmamba}
\end{figure*}

The original Mamba block~\citep{gu2023mamba} integrates the H3 block~\citep{fu2022hungry} with a gated MLP, where the H3 block uses a 1D-conv before the SSM layer to capture local information within nearby tokens, as illustrated in Figure~\ref{fig:cdmamba2}. 
However, since channels in TS do not have an inherent sequential order, 
we eliminate the 1D-conv from the Mamba block, resulting in the proposed CD-Mamba block.
Figure~\ref{fig:cdmamba} shows the overall architecture of the proposed CD-Mamba block, where the 1D-conv before the selective SSM is removed from the original Mamba block~\citep{gu2023mamba}. 

\newpage
\section{Full Results of Time Series Forecasting} 
\label{sec:full}
Table~\ref{tbl:full_fcst} shows the full results of TS forecasting tasks across four different horizons, highlighting the effectiveness of our method.

\begin{table*}[h]
\vspace{20pt}
\centering
\begin{adjustbox}{max width=.999\textwidth}
\begin{NiceTabular}{c|c|cccc|cc|cc|cc|cc|cc|cc|cc|cc}
\toprule
\multicolumn{2}{c}{\multirow{2.5}{*}{Models}} & \multicolumn{4}{c}{SOR-Mamba} & \multicolumn{2}{c}{\multirow{2.5}{*}{S-Mamba}} & \multicolumn{2}{c}{\multirow{2.5}{*}{iTransformer}} & \multicolumn{2}{c}{\multirow{2.5}{*}{RLinear}} & \multicolumn{2}{c}{\multirow{2.5}{*}{PatchTST}} & \multicolumn{2}{c}{\multirow{2.5}{*}{Crossformer}} & \multicolumn{2}{c}{\multirow{2.5}{*}{TiDE}} & \multicolumn{2}{c}{\multirow{2.5}{*}{TimesNet}} & \multicolumn{2}{c}{\multirow{2.5}{*}{DLinear}} \\
\cmidrule(lr){3-6}
\multicolumn{2}{c}{}&\multicolumn{2}{c}{FT}&\multicolumn{2}{c}{SL}& \\
\cmidrule(lr){1-2}\cmidrule(lr){3-4}\cmidrule(lr){5-6}\cmidrule(lr){7-8} \cmidrule(lr){9-10}\cmidrule(lr){11-12}\cmidrule(lr){13-14}\cmidrule(lr){15-16}\cmidrule(lr){17-18}\cmidrule(lr){19-20}\cmidrule(lr){21-22}
\multicolumn{2}{c}{Metric} & MSE & MAE & MSE & MAE & MSE & MAE & MSE & MAE & MSE & MAE & MSE & MAE & MSE & MAE & MSE & MAE & MSE & MAE & MSE & MAE \\
\toprule

\multirow{5.5}{*}{\rotatebox{90}{ETTh1}} 
&  96 & \first{.377}&\second{.398}&\second{.385}&\second{.398}&\second{.385} & .404 & .387 & .405 & .386 & \first{.395} & .414 & .419 & .423 & .448 & .479 & .464 & \fourth{.384} & .402 & .386 & \fourth{.400} \\

& 192 & \first{.428}&{.429}&\second{.435}&\second{.428}&.445 & .441 & .441 & .436 & {.437} & \first{.424} & .460 & .445 & .471 & .474 & .525 & .492 & \fourth{.436} & \fourth{.429} & .437 & .432 \\

& 336 & \first{.464}&\second{.448}&\second{.474}&\second{.448}&.491 & .462 & .487 & \fourth{.458} & \fourth{.479} & \first{.446} & .501 & .466 & .570 & .546 & .565 & .515 & .491 & .469 & .481 & .459 \\

& 720 & \first{.464}&\first{.469}&\second{.478}&{.471}&.506 & .497 & .509 & .494 & \third{.481} & \second{.470} & \fourth{.500} & \fourth{.488} & .653 & .621 & .594 & .558 & .521 & .500 & .519 & .516\\
\cmidrule(lr){2-22}

& \cellcolor{gray!20} Avg. & \cellcolor{gray!20} \first{.433}&\cellcolor{gray!20}\second{.436}&\cellcolor{gray!20}\second{.442}&\cellcolor{gray!20}{.438}&\cellcolor{gray!20}.457 &\cellcolor{gray!20} .452 &\cellcolor{gray!20} .457 &\cellcolor{gray!20} \fourth{.449} &\cellcolor{gray!20} \fourth{.446} &\cellcolor{gray!20} \first{.434} &\cellcolor{gray!20} .469 &\cellcolor{gray!20} .454 &\cellcolor{gray!20} .529 &\cellcolor{gray!20} .522 &\cellcolor{gray!20} .541 &\cellcolor{gray!20} .507 &\cellcolor{gray!20} .458 &\cellcolor{gray!20} .450 &\cellcolor{gray!20} .456 &\cellcolor{gray!20} .452 \\
\midrule

\multirow{5.5}{*}{\rotatebox{90}{ETTh2}} 
& 96  & \second{.292}&\second{.348}&.299&\second{.348}&\fourth{.297} & \fourth{.349} & .301 & .350 & \first{.288} & \first{.338} & .302 & \second{.348} & .745 & .584 & .400 & .440 & .340 & .374 & .333 & .387\\

& 192 & \first{.372}&\second{.397}&.375&.399&\fourth{.378} & \fourth{.399} & .381 & .399 & \second{.374} & \first{.390} & .388 & .400 & .877 & .656 & .528 & .509 & .402 & .414 & .477 & .476 \\

& 336 & \first{.415}&\second{.431}&\second{.423}&{.435}&\fourth{.425} & \fourth{.435} & .427 & .434 & \first{.415} & \first{.426} & .426 & .433 & 1.043 & .731 & .643 & .571 & .452 & .452 & .594 & .541 \\

& 720 & \second{.423}&\second{.445}&.431&.446&\fourth{.432} & \fourth{.448} & .430 & .446 & \first{.420} & \first{.440} & .431 & .446 & 1.104 & .763 & .874 & .679 & .462 & .468 & .831 & .657 \\
\cmidrule(lr){2-22}

& \cellcolor{gray!20} Avg. &\cellcolor{gray!20} \second{.376}&\cellcolor{gray!20}\second{.405}&\cellcolor{gray!20}.382&\cellcolor{gray!20}.407&\cellcolor{gray!20}\fourth{.383} &\cellcolor{gray!20} \fourth{.408} &\cellcolor{gray!20} .384 &\cellcolor{gray!20} .407 &\cellcolor{gray!20} \first{.374} &\cellcolor{gray!20} \first{.398} &\cellcolor{gray!20} .387 &\cellcolor{gray!20} .407 &\cellcolor{gray!20} .942 &\cellcolor{gray!20} .684 &\cellcolor{gray!20} .611 &\cellcolor{gray!20} .550 &\cellcolor{gray!20} .414 &\cellcolor{gray!20} .427 &\cellcolor{gray!20} .559 &\cellcolor{gray!20} .515 \\
\midrule

\multirow{5.5}{*}{\rotatebox{90}{ETTm1}} 
&  96 & \first{.324}&\first{.362}&\second{.326}&\second{.367} &\second{.326} & \fourth{.368} & .342 & \fourth{.377} & .355 & .376 & \fourth{.329} & \second{.367} & .404 & .426 & .364 & .387 & .338 & .375 & .345 & .372 \\

& 192 & \second{.369}&\first{.385}&\fourth{.375}&\second{.387}&.378 & .393 & .383 & .396 & .391 & .392 & \first{.367} & \first{.385} & .450 & .451 & .398 & .404 & {.374} & \second{.387} & .380 & .389 \\

& 336 & \second{.402}&\first{.408}&.408&\first{.408}&.410 & .414 & .418 & .418 & .424 & .415 & \first{.399} & \second{.410} & .532 & .515 & .428 & .425 & \fourth{.410} & \fourth{.411} & .413 & .413 \\

& 720 & \second{.467}&\second{.444}&.472&\second{.444}&.474 & \fourth{.451} & .487 & .456 & .487 & .450 & \first{.454} & \first{.439} & .666 & .589 & .487 & .461 & .478 & .450 & \fourth{.474} & .453\\
\cmidrule(lr){2-22}

&\cellcolor{gray!20} Avg. &\cellcolor{gray!20} \second{.391}&\cellcolor{gray!20}\first{.400}&\cellcolor{gray!20}.396&\cellcolor{gray!20}\second{.401}&\cellcolor{gray!20}\fourth{.398} &\cellcolor{gray!20} \fourth{.407} &\cellcolor{gray!20} .408 &\cellcolor{gray!20} .412 &\cellcolor{gray!20} .414 &\cellcolor{gray!20} .407 &\cellcolor{gray!20} \first{.387} &\cellcolor{gray!20} \first{.400} &\cellcolor{gray!20} .513 &\cellcolor{gray!20} .496 &\cellcolor{gray!20} .419 &\cellcolor{gray!20} .419 &\cellcolor{gray!20} .400 &\cellcolor{gray!20} .406 &\cellcolor{gray!20} .403 &\cellcolor{gray!20} .407 \\
\midrule

\multirow{5.5}{*}{\rotatebox{90}{ETTm2}} 
& 96  & \second{.179}&\second{.261}&.181&.265&\fourth{.182} & \fourth{.266} & .186 & .272 & .182 & .265 & \first{.175} & \first{.259} & .287 & .366 & .207 & .305 & .187 & .267 & .193 & .292 \\

& 192 & \first{.241}&\second{.304}&\second{.246}&.307&.252 & .313 & .254 & .314 & \second{.246} & \second{.304} & \first{.241} & \first{.302} & .414 & .492 & .290 & .364 & .249 & .309 & .284 & .362 \\

& 336 & \first{.302}&\first{.342}&.306&.345&.313 & .349 & .317 & .353 & \fourth{.307} & \first{.342} & \first{.305} & \second{.343} & .597 & .542 & .377 & .422 & .321 & .351 & .369 & .427 \\

& 720 & \first{.401}&\second{.400}&.403&\third{.401}&.416 & .409 & .412 & .407 & \fourth{.407} & \first{.398} & \second{.402} & \second{.400} & 1.730 & 1.042 & .558 & .524 & .408 & .403 & .554 & .522 \\
\cmidrule(lr){2-22}

&\cellcolor{gray!20} Avg. &\cellcolor{gray!20} \first{.281}&\cellcolor{gray!20}\second{.327}&\cellcolor{gray!20}\second{.284}&\cellcolor{gray!20}.329&\cellcolor{gray!20}.290 &\cellcolor{gray!20} .333 &\cellcolor{gray!20} .293 &\cellcolor{gray!20} .337 &\cellcolor{gray!20} \fourth{.286} &\cellcolor{gray!20} \second{.327} &\cellcolor{gray!20} \first{.281} &\cellcolor{gray!20} \first{.326} &\cellcolor{gray!20} .757 &\cellcolor{gray!20} .610 &\cellcolor{gray!20} .358 &\cellcolor{gray!20} .404 &\cellcolor{gray!20} .291 &\cellcolor{gray!20} .333 &\cellcolor{gray!20} .350 &\cellcolor{gray!20} .401\\

\midrule

\multirow{5.5}{*}{\rotatebox{90}{PEMS03}} 
& 12 &\first{.066}&\first{.170}&\first{.066}&\first{.170}& \first{.066} & \second{.171} & \second{.071} & \fourth{.174} & .126 & .236 & .099 & .216 & .090 & .203 & .178 & .305 & .085 & .192 & .122 & .243 \\

& 24 &\first{.088}&\first{.197}&\second{.090}&\second{.200}& \first{.088} & \first{.197} & \fourth{.097} & \fourth{.208} & .246 & .334 & .142 & .259 & .121 & .240 & .257 & .371 & .118 & .223 & .201 & .317 \\

& 48 &\first{.134}&\first{.245}&.167&.280& \fourth{.165} & \fourth{.277} & \second{.161} & \second{.272}   & .551 & .529 & .211 & .319 & .202 & .317 & .379 & .463 & .155 & .260 & .333 & .425 \\

& 96 &\first{.193}&\first{.297}&.225&.318& \second{.213} & \second{.313} & {.240} & {.338} & 1.057 & .787 & .269 & .370 & .262 & .367 & .490 & .539 & .228 & .317 & .457 & .515 \\
\cmidrule(lr){2-22}

&\cellcolor{gray!20} Avg. &\cellcolor{gray!20}\first{.121}&\cellcolor{gray!20}\first{.227}&\cellcolor{gray!20}.137&\cellcolor{gray!20}.242&\cellcolor{gray!20} \second{.133} &\cellcolor{gray!20} \second{.240} &\cellcolor{gray!20} {.142} &\cellcolor{gray!20} {.248} &\cellcolor{gray!20} .495 &\cellcolor{gray!20} .472 &\cellcolor{gray!20} .180 &\cellcolor{gray!20} .291 &\cellcolor{gray!20} .169 &\cellcolor{gray!20} .281 &\cellcolor{gray!20} .326 &\cellcolor{gray!20} .419 &\cellcolor{gray!20} .147 &\cellcolor{gray!20} .248 &\cellcolor{gray!20} .278 &\cellcolor{gray!20} .375\\
\midrule

\multirow{5.5}{*}{\rotatebox{90}{PEMS04}} 
& 12 &\second{.074}&\first{.175}&.077&.180& \first{.073} & \second{.177} & \fourth{.081} & \fourth{.188} & .138 & .252 & .105 & .224 & .098 & .218 & .219 & .340 & .087 & .195 & .148 & .272 \\

& 24 &\second{.086}&\first{.192}&.091&\second{.197}& \first{.084} & \first{.192} & \fourth{.099} & \fourth{.211} & .258 & .348 & .153 & .275 & .131 & .256 & .292 & .398 & .103 & .215 & .224 & .340 \\

& 48 &\second{.106}&\second{.214}&.115&.221& \first{.101} & \first{.213} & \fourth{.133} & \fourth{.246} & .572 & .544 & .229 & .339 & .205 & .326 & .409 & .478 & .136 & .250 & .355 & .437 \\

& 96 &\second{.129}&\first{.233}&.143&.248& \first{.125} & \second{.236} & \fourth{.172} & \fourth{.283} & 1.137 & .820 & .291 & .389 & .402 & .457 & .492 & .532 & .190 & .303 & .452 & .504 \\
\cmidrule(lr){2-22}

&\cellcolor{gray!20} Avg. &\cellcolor{gray!20}\second{.099}&\cellcolor{gray!20}\first{.203}&\cellcolor{gray!20}.107&\cellcolor{gray!20}.212&\cellcolor{gray!20} \first{.096} &\cellcolor{gray!20} \second{.205} &\cellcolor{gray!20} \fourth{.121} &\cellcolor{gray!20} \fourth{.232} &\cellcolor{gray!20} .526 &\cellcolor{gray!20} .491 &\cellcolor{gray!20} .195 &\cellcolor{gray!20} .307 &\cellcolor{gray!20} .209 &\cellcolor{gray!20} .314 &\cellcolor{gray!20} .353 &\cellcolor{gray!20} .437 &\cellcolor{gray!20} .129 &\cellcolor{gray!20} .241 &\cellcolor{gray!20} .295 &\cellcolor{gray!20} .388 \\
\midrule

\multirow{5.5}{*}{\rotatebox{90}{PEMS07}} 
& 12 &\first{.059}&\first{.155}&\second{.060}&\second{.156}& \second{.060} & \third{.157} & \fourth{.067} & \fourth{.165} & .118 & .235 & .095 & .207 & .094 & .200 & .173 & .304 & .082 & .181 & .115 & .242\\

& 24 &\first{.076}&\first{.174}&\second{.082}&\second{.182}& \second{.082} & \third{.184} & \fourth{.088} & \fourth{.190} & .242 & .341 & .150 & .262 & .139 & .247 & .271 & .383 & .101 & .204 & .210 & .329 \\

& 48 &\first{.098}&\first{.199}&.107&.209& \second{.100} & \second{.204} & \fourth{.113} & \fourth{.218} & .562 & .541 & .253 & .340 & .311 & .369 & .446 & .495 & .134 & .238 & .398 & .458 \\

& 96 &\first{.117}&\first{.218}&\first{.117}&\first{.218}& \first{.117} & \first{.218} & \second{.172} & \second{.283} & 1.096 & .795 & .346 & .404 & .396 & .442 & .628 & .577 & .181 & .279 & .594 & .553 \\
\cmidrule(lr){2-22}

&\cellcolor{gray!20}  Avg. &\cellcolor{gray!20}\first{.088}&\cellcolor{gray!20}\first{.186}&\cellcolor{gray!20} .091 &\cellcolor{gray!20}\second{.191}&\cellcolor{gray!20} \second{.090} &\cellcolor{gray!20} \second{.191} &\cellcolor{gray!20} \fourth{.102} &\cellcolor{gray!20} \fourth{.205} &\cellcolor{gray!20} .504 &\cellcolor{gray!20} .478 &\cellcolor{gray!20} .211 &\cellcolor{gray!20} .303 &\cellcolor{gray!20} .235 &\cellcolor{gray!20} .315 &\cellcolor{gray!20} .380 &\cellcolor{gray!20} .440 &\cellcolor{gray!20} .124 &\cellcolor{gray!20} .225 &\cellcolor{gray!20} .329 &\cellcolor{gray!20} .395 \\
\midrule

\multirow{5.5}{*}{\rotatebox{90}{PEMS08}}
& 12&\second{.078}&\second{.178}&\first{.076}&\first{.176} & \first{.076} & \second{.178} & \fourth{.088} & \fourth{.193} & .133 & .247 & .168 & .232 & .165 & .214 & .227 & .343 & .112 & .212 & .154 & .276 \\

& 24 &\first{.103}&\first{.205}&\second{.109}&\second{.212}& .110 & \third{.216} & \fourth{.138} & \fourth{.243} & .249 & .343 & .224 & .281 & .215 & .260 & .318 & .409 & .141 & .238 & .248 & .353 \\

& 48 &\first{.159}&\first{.250}&\second{.172}&.264& \third{.173} & \second{.254} & \fourth{.334} & \fourth{.353} & .569 & .544 & .321 & .354 & .315 & .355 & .497 & .510 & .198 & .283 & .440 & .470\\

& 96 &\first{.229}&\first{.295}&.290&.334& \second{.271} & \second{.321} & {.458} & {.436} & 1.166 & .814 & .408 & .417 & .377 & .397 & .721 & .592 & .320 & .351 & .674 & .565 \\
\cmidrule(lr){2-22}

&\cellcolor{gray!20} Avg. &\cellcolor{gray!20}\first{.142}&\cellcolor{gray!20}\first{.232}&\cellcolor{gray!20}\third{.162}&\cellcolor{gray!20}.247&\cellcolor{gray!20} \second{.157} &\cellcolor{gray!20} \second{.242} &\cellcolor{gray!20} {.254} &\cellcolor{gray!20} {.306} &\cellcolor{gray!20} .529 &\cellcolor{gray!20} .487 &\cellcolor{gray!20} .280 &\cellcolor{gray!20} .321 &\cellcolor{gray!20} .268 &\cellcolor{gray!20} .307 &\cellcolor{gray!20} .441 &\cellcolor{gray!20} .464 &\cellcolor{gray!20} .193 &\cellcolor{gray!20} .271 &\cellcolor{gray!20} .379 &\cellcolor{gray!20} .416 \\

\midrule

\multirow{5.5}{*}{\rotatebox{90}{Exchange}} 
& 96  & \first{.085}&\first{.204}&\first{.085}&\second{.205}&\second{.086} & .206 & \second{.086} & \fourth{.206} & .093 & .217 & \fourth{.088} & \second{.205} & .256 & .367 & .094 & .218 & .107 & .234 & \fourth{.088} & .218 \\

& 192 & \third{.179}&\second{.301}&\third{.179}&\second{.301}&.181 & .303 & \second{.177} & \first{.299} & .184 & .307 & \first{.176} & \first{.299} & .470 & .509 & .184 & .307 & .226 & .344 & \first{.176} & .315\\

& 336 & .329&\second{.415}&.331&.417&.331 & .417 & .338 & \fourth{.422} & .351 & .432 & \first{.301} & \first{.397} & 1.268 & .883 & .349 & .431 & .367 & .448 & \second{.313} & .427\\

& 720 & \first{.838}&\second{.690}&.860&.698&.858 & \first{.599} & \fourth{.847} & \third{.691} & .886 & .714 & .901 & .714 & 1.767 & 1.068 & .852 & .698 & .964 & .746 & \second{.839} & \fourth{.695}\\
\cmidrule(lr){2-22}

&\cellcolor{gray!20} Avg. &\cellcolor{gray!20} \second{.358}&\cellcolor{gray!20}\first{.402}&\cellcolor{gray!20}.363&\cellcolor{gray!20}{.405}&\cellcolor{gray!20}.364 &\cellcolor{gray!20} .407 &\cellcolor{gray!20} \fourth{.368} &\cellcolor{gray!20} {.409} &\cellcolor{gray!20} .378 &\cellcolor{gray!20} .417 &\cellcolor{gray!20} .367 &\cellcolor{gray!20} \second{.404} &\cellcolor{gray!20} .940 &\cellcolor{gray!20} .707 &\cellcolor{gray!20} .370 &\cellcolor{gray!20} .413 &\cellcolor{gray!20} .416 &\cellcolor{gray!20} .443 &\cellcolor{gray!20} \first{.354} &\cellcolor{gray!20} .414\\
\midrule

\multirow{5.5}{*}{\rotatebox{90}{Weather}} 
& 96  & {.174}&\second{.212}&.175&.215&\second{.165} & \first{.209} & .174 & \fourth{.215} & .192 & .232 & .177 & .218 & \first{.158} & .230 & .202 & .261 & .172 & .220 & .196 & .255 \\

& 192 & {.221}&\first{.255}&{.221}&\first{.255}&\second{.215} & \first{.255} & .224 & \second{.258} & .240 & .271 & .225 & .259 & \first{.206} & .277 & .242 & .298 & {.219} & .261 & .237 & .296 \\

& 336 & \second{.277}&\first{.295}&\second{.277}&\second{.296}&\first{.273} & \second{.296} & .281 & \third{.298} & .292 & .307 & .278 & \fourth{.297} & \first{.273} & .335 & .287 & .335 & .280 & .306 & .283 & .335 \\

& 720 & \second{.353}&\first{.348}&.355&\first{.348}&\second{.353} & \second{.349} & .359 & {.351} & .364 & .353 & .354 & \first{.348} & .398 & .418 & .351 & .386 & .365 & .359 & \first{.345} & .381 \\
\cmidrule(lr){2-22}

&\cellcolor{gray!20} Avg. &\cellcolor{gray!20} \second{.256}&\cellcolor{gray!20}\first{.277}&\cellcolor{gray!20}.257&\cellcolor{gray!20}\second{.278}&\cellcolor{gray!20}\first{.252} &\cellcolor{gray!20} \first{.277} &\cellcolor{gray!20} \fourth{.260} &\cellcolor{gray!20} {.281} &\cellcolor{gray!20} .272 &\cellcolor{gray!20} .291 &\cellcolor{gray!20} .259 &\cellcolor{gray!20} .281 &\cellcolor{gray!20} .259 &\cellcolor{gray!20} .315 &\cellcolor{gray!20} .271 &\cellcolor{gray!20} .320 &\cellcolor{gray!20} .259 &\cellcolor{gray!20} .287 &\cellcolor{gray!20} .265 &\cellcolor{gray!20} .317 \\
\midrule

\multirow{5.5}{*}{\rotatebox{90}{Solar}} 
& 96  & \first{.194}&\first{.229}&.207&.246&\fourth{.207} & \fourth{.246} & \second{.201} & \second{.234} & .322 & .339 & .234 & .286 & .310 & .331 & .312 & .399 & .250 & .292 & .290 & .378 \\

& 192 & \first{.228}&\first{.256}&.239&.270&\fourth{.240} & \fourth{.272} & \second{.238} & \second{.261} & .359 & .356 & .267 & .310 & .734 & .725 & .339 & .416 & .296 & .318 & .320 & .398 \\

& 336 & \first{.247}&\second{.276}&.260&.287&\fourth{.262} & \fourth{.290} & \second{.248} & \first{.273} & .397 & .369 & .290 & .315 & .750 & .735 & .368 & .430 & .319 & .330 & .353 & .415 \\

& 720 & \second{.251}&\first{.275}&.264&\second{.291}&\fourth{.267} & \fourth{.293} & \first{.249} & \first{.275} & .397 & .356 & .289 & .317 & .769 & .765 & .370 & .425 & .338 & .337 & .356 & .413 \\
\cmidrule(lr){2-22}

&\cellcolor{gray!20} Avg. &\cellcolor{gray!20} \first{.230}&\cellcolor{gray!20}\first{.259}&\cellcolor{gray!20}.242&\cellcolor{gray!20}.274&\cellcolor{gray!20}\fourth{.244} &\cellcolor{gray!20} \fourth{.275} &\cellcolor{gray!20} \second{.234} &\cellcolor{gray!20} \second{.261} &\cellcolor{gray!20} .369 &\cellcolor{gray!20} .356 &\cellcolor{gray!20} .270 &\cellcolor{gray!20} .307 &\cellcolor{gray!20} .641 &\cellcolor{gray!20} .639 &\cellcolor{gray!20} .347 &\cellcolor{gray!20} .417 &\cellcolor{gray!20} .301 &\cellcolor{gray!20} .319 &\cellcolor{gray!20} .330 &\cellcolor{gray!20} .401 \\
\midrule

\multirow{5.5}{*}{\rotatebox{90}{ECL}} 
& 96  & \first{.139}&\second{.235}&\first{.139}&\first{.233}&\first{.139} & \third{.237} & \second{.148} & \fourth{.240} & .201 & .281 & .181 & .270 & .219 & .314 & .237 & .329 & .168 & .272 & .197 & .282 \\

& 192 & \second{.160}&\second{.254}&\first{.158}&\first{.249}&\third{.165} & \fourth{.261} & \fourth{.167} & \third{.258} & .201 & .283 & .188 & .274 & .231 & .322 & .236 & .330 & .184 & .289 & .196 & .285 \\

& 336 & \first{.176}&\first{.271}&\second{.177}&\first{.271}&\second{.177} & \fourth{.274} & \fourth{.179} & \second{.272} & .215 & .298 & .204 & .293 & .246 & .337 & .249 & .344 & .198 & .300 & .209 & .301 \\

& 720 & \first{.198}&\first{.292}&\second{.201}&\second{.293}&\third{.214} & \third{.304} & .220 & \fourth{.310} & .257 & .331 & .246 & .324 & .280 & .363 & .284 & .373 & \fourth{.220} & .320 & .245 & .333 \\
\cmidrule(lr){2-22}

&\cellcolor{gray!20} Avg. &\cellcolor{gray!20} \first{.168}&\cellcolor{gray!20}\second{.264}&\cellcolor{gray!20}\second{.169}&\cellcolor{gray!20}\first{.262}&\cellcolor{gray!20}\third{.174} &\cellcolor{gray!20} \third{.269} &\cellcolor{gray!20} \fourth{.179} &\cellcolor{gray!20} \fourth{.270} &\cellcolor{gray!20} .219 &\cellcolor{gray!20} .298 &\cellcolor{gray!20} .205 &\cellcolor{gray!20} .290 &\cellcolor{gray!20} .244 &\cellcolor{gray!20} .334 &\cellcolor{gray!20} .251 &\cellcolor{gray!20} .344 &\cellcolor{gray!20} .192 &\cellcolor{gray!20} .295 &\cellcolor{gray!20} .212 &\cellcolor{gray!20} .300 \\
\midrule

\multirow{5.5}{*}{\rotatebox{90}{Traffic}} 
& 96 &\first{.378}&{.261}&\first{.378}&\first{.259} & \second{.379} & \second{.260} & \fourth{.395} & \fourth{.268} & .649 & .389 & .462 & .295 & .522 & .290 & .805 & .493 & .593 & .321 & .650 & .396 \\

& 192 &\first{.393}&\first{.269}&\second{.399}&\second{.270}& \third{.409} & \third{.272} & \fourth{.417} & \fourth{.277} & .601 & .366 & .466 & .296 & .530 & .293 & .756 & .474 & .617 & .336 & .598 & .370 \\

& 336 &\first{.399}&\first{.272}&\second{.416}&.279& \third{.418} & \second{.277} & \fourth{.433} & \fourth{.283} & .609 & .369 & .482 & .304 & .558 & .305 & .762 & .477 & .629 & .336 & .605 & .373\\

& 720 &\first{.437}&\first{.290}&\second{.456}&\second{.297}& \third{.461} & \second{.297} & \fourth{.467} & \fourth{.300} & .647 & .387 & .514 & .322 & .589 & .328 & .719 & .449 & .640 & .350 & .645 & .394 \\
\cmidrule(lr){2-22}

&\cellcolor{gray!20} Avg. &\cellcolor{gray!20}\first{.402}&\cellcolor{gray!20}\first{.273}&\cellcolor{gray!20}\second{.412}&\cellcolor{gray!20}\second{.276}&\cellcolor{gray!20} \third{.417} &\cellcolor{gray!20} \third{.277} &\cellcolor{gray!20} \fourth{.428} &\cellcolor{gray!20} \fourth{.282} &\cellcolor{gray!20} .626 &\cellcolor{gray!20} .378 &\cellcolor{gray!20} .481 &\cellcolor{gray!20} .304 &\cellcolor{gray!20} .550 &\cellcolor{gray!20} .304 &\cellcolor{gray!20} .760 &\cellcolor{gray!20} .473 &\cellcolor{gray!20} .620 &\cellcolor{gray!20} .336 &\cellcolor{gray!20} .625 &\cellcolor{gray!20} .383 \\
\midrule
\rowcolor{gray!20} \multicolumn{2}{c}{$1^{\text{st}}$ Count}  &  33 & 31 & 7 & 10 & 10 & 7 & 1 & 3 & 3& 9& 8& 7& 3 & 0 & 0 & 0 & 0 & 0 & 2 & 0 \\
\rowcolor{gray!20} \multicolumn{2}{c}{$2^{\text{nd}}$ Count}& 15 & 19 & 18 & 19 & 13 & 13 & 9 & 6 & 2 & 2 & 1 & 6 & 0 & 0 & 0 & 0 & 0 & 1 & 2 & 0 \\
\bottomrule
\end{NiceTabular}
\end{adjustbox}
\caption{Full results of TS forecasting tasks.}
\label{tbl:full_fcst}
\end{table*}

\newpage
\section{Ablation Study}
\label{sec:ablation}

To demonstrate the effectiveness of our method, we conduct an ablation study using four ETT datasets~\citep{zhou2021informer} to assess the impact of the following components, where the results are shown in Table~\ref{fig:ablation_full}.
The results indicate that incorporating all components yields the best performance, and adding the regularization term enhances the performance even with the bidirectional Mamba.

\begin{table*}[h]
\vspace{20pt}
\centering
\begin{adjustbox}{max width=.999\textwidth}
\begin{NiceTabular}{c|cccc|cccc|c}
\toprule
\multirow{2.5}{*}{Method} & \multicolumn{2}{c}{Mamba} & \multirow{2.5}{*}{Reg.} & \multirow{2.5}{*}{CCM} & \multirow{2.5}{*}{ETTh1} & \multirow{2.5}{*}{ETTh2} & \multirow{2.5}{*}{ETTm1} & \multirow{2.5}{*}{ETTm2} & \multirow{2.5}{*}{Avg.} \\
\cmidrule(lr){2-3}
& \# & w/o conv. & & & & & \\
\midrule
S-Mamba & Bi & -& - & - & .457 & .383  & .398 & .290 & .382 \\
- & Bi & \cmark &- & - & {.441} & .383  & .396 & .285 & .376\\
- & Bi & - & \cmark &  & .452 & .382  & {.394} & .286 & .378 \\
- & Bi & \cmark & \cmark &  & .443 & \second{.381}  & .393 & .285 & \second{.376} \\
- & Bi & \cmark & \cmark & \cmark & \second{.435} & \first{.376}  & \first{.390} & \first{.281} & \first{.370} \\
\midrule
- & Uni & - & - & - & .455 & .383  & .403 & .289 & .383 \\
- & Uni & \cmark &- & - & .442 & {.382}  & .400 & .285 & .377 \\
- & Uni & - & \cmark & - & .449 & {.382}  & .396 & .285 & .378 \\
- & Uni & \cmark & \cmark &- & .442 & {.382} & {.396} & \second{.284} & \second{.376} \\
SOR-Mamba & Uni & \cmark & \cmark & \cmark & \first{.433} & \first{.376} & \second{.391} & \first{.281}  & \first{.370}\\
\bottomrule
\end{NiceTabular}
\end{adjustbox}
\caption{Ablation studies with four ETT datasets.}
\label{fig:ablation_full}
\end{table*}

\vspace{30pt}
\section{Channel Orders for Two Views}
\label{sec:fixed_vs_random}

\begin{figure*}[h]
\vspace{20pt}
\centering
\includegraphics[width=.999\textwidth]{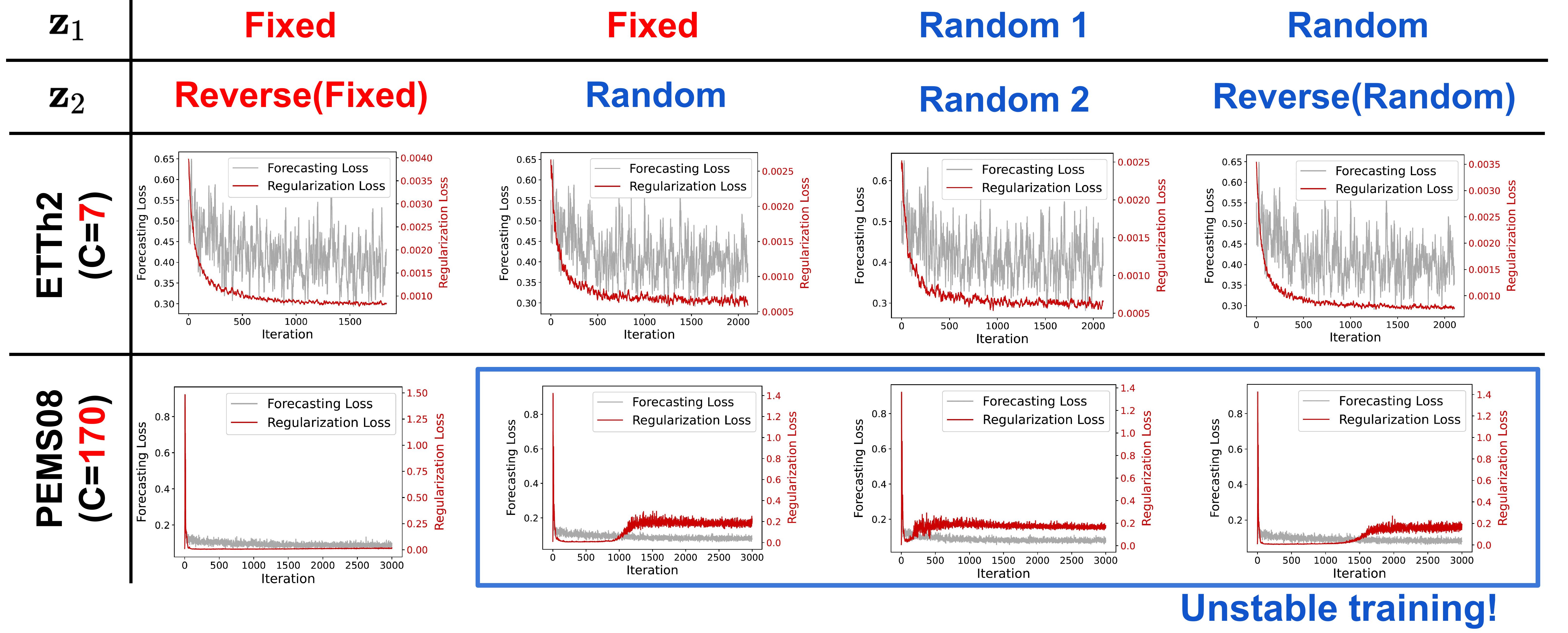} 
\caption{Fixed vs. random order for generating two views, $\mathbf{z}_1$ and $\mathbf{z}_2$.}
\label{fig:train_loss}
\end{figure*}

Figure~\ref{fig:train_loss} illustrates the four candidates for generating two embedding vectors, $\mathbf{z}_1$ and $\mathbf{z}_2$, for regularization, based on whether the channel order is fixed or randomly permuted in each iteration. 
Results in Table~\ref{tbl:fixedvsrandom} indicate that fixing the order during training yields the best performance, with performance degrading as the order becomes random, especially with many channels, though it remains robust with fewer channels.
We argue that a fixed order is preferable due to the instability introduced by randomness during training, as shown in Figure~\ref{fig:train_loss}, which displays the training loss for two datasets~\citep{zhou2021informer,liu2022scinet} with varying numbers of channels.
The figure indicates that a random order causes instability, particularly with the regularization loss.

\newpage
\section{Robustness to Channel Order}
\label{sec:channel_order_robust}
To demonstrate that the proposed method effectively addresses the sequential order bias, 
we evaluate performance variations by permuting the channel order
with five datasets~\citep{zhou2021informer,wu2021autoformer}.
Table~\ref{tbl:robust_full} shows the results,
which indicate a small standard deviation across all horizons.

\begin{table*}[h]
\vspace{20pt}
\centering
\begin{adjustbox}{max width=0.999\textwidth}
\begin{NiceTabular}{c|ccccc}
\toprule
$H$ & ETTh1 & ETTh2 & ETTm1 & ETTm2 & Exchange\\
\midrule
96 & $.377_{\pm \textbf{\textcolor{red}{.0003}}}$ & $.292_{\pm \textbf{\textcolor{red}{.0011}}}$ & $.324_{\pm \textbf{\textcolor{red}{.0005}}}$ & $.179_{\pm \textbf{\textcolor{red}{.0003}}}$  & $.085_{\pm \textbf{\textcolor{red}{.0001}}}$ \\
192 & $.428_{\pm \textbf{\textcolor{red}{.0002}}}$ & $.372_{\pm \textbf{\textcolor{red}{.0000}}}$ & $.369_{\pm \textbf{\textcolor{red}{.0005}}}$ & $.241_{\pm \textbf{\textcolor{red}{.0002}}}$  & $.179_{\pm \textbf{\textcolor{red}{.0001}}}$ \\
336 & $.464_{\pm \textbf{\textcolor{red}{.0002}}}$ & $.415_{\pm \textbf{\textcolor{red}{.0002}}}$ & $.402_{\pm \textbf{\textcolor{red}{.0003}}}$ & $.302_{\pm \textbf{\textcolor{red}{.0001}}}$  & $.329_{\pm \textbf{\textcolor{red}{.0002}}}$ \\
720 & $.464_{\pm \textbf{\textcolor{red}{.0004}}}$ & $.423_{\pm \textbf{\textcolor{red}{.0001}}}$ & $.467_{\pm \textbf{\textcolor{red}{.0009}}}$ & $.401_{\pm \textbf{\textcolor{red}{.0001}}}$  & $.838_{\pm \textbf{\textcolor{red}{.0014}}}$ \\
\midrule
Avg. & $.434_{\pm \textbf{\textcolor{red}{.0002}}}$ & $.423_{\pm \textbf{\textcolor{red}{.0003}}}$ & $.391_{\pm \textbf{\textcolor{red}{.0001}}}$ & $.281_{\pm \textbf{\textcolor{red}{.0001}}}$  & $.358_{\pm \textbf{\textcolor{red}{.0003}}}$  \\
\bottomrule
\end{NiceTabular}
\end{adjustbox}
\caption{Robustness to channel order.}
\label{tbl:robust_full}
\end{table*}

\vspace{30pt}
\section{Robustness to Hyperparameter \texorpdfstring{$\lambda$}{}}
\label{sec:robust_lambda}
Table~\ref{tbl:robust} shows the average MSE across four different horizons for the four ETT datasets~\citep{zhou2021informer}, using various values of $\lambda$ that control the contribution of the regularization term. The results demonstrate the effectiveness of the regularization and its robustness to $\lambda$.

\begin{table*}[h]
\centering
\vspace{10pt}
\begin{adjustbox}{max width=1.00\textwidth}
\begin{NiceTabular}{c|c|cccc|c}
\toprule
\multirow{4}{*}{Dataset} & \multicolumn{5}{c}{SOR-Mamba} & \multirow{4}{*}{S-Mamba} \\
\cmidrule{2-6}
& w/o Reg. & \multicolumn{4}{c}{w/ Reg.} & \\
\cmidrule(lr){2-2} \cmidrule(lr){3-6}
& 0& 0.001& 0.01& 0.1& 0.2 & \\
\midrule
ETTh1 & \second{.439} & \first{.433}& \first{.433}& \first{.433}& \first{.433} & .457 \\
ETTh2 & \second{.382} & \first{.376}& \first{.376}& \first{.376}& \first{.376} & .383 \\
ETTm1 & .403 & \first{.391}& \first{.391}& \first{.391}& \first{.391} & \second{.398} \\
ETTm2 & \second{.285} & \first{.281}& \first{.281}& \first{.281}& \first{.281} & .290 \\
\bottomrule
\end{NiceTabular}
\end{adjustbox}
\caption{Robustness to choice of $\lambda$ for regularization.}
 \label{tbl:robust}
\end{table*}

\newpage
\section{Robustness to Distance Metric}
\label{sec:robust_distance}

To assess whether SOR-Mamba is sensitive to the choice of distance metric $d$ for the regularization term and CCM when comparing the two matrices, we compare various metrics, including (negative) cosine similarity,  $\ell_1$ loss, and  $\ell_2$ loss. Tables~\ref{tbl:robust_distance_reg} and \ref{tbl:robust_distance_ccm} show the average MSE across four different horizons for the distance metric used in the regularization term and CCM, respectively, demonstrating that the performance is robust to the choice of distance metric, where we choose $\ell_2$ loss throughout the experiment for both metrics.

\vspace{15pt}
\begin{figure*}[h]
    \centering
    \begin{minipage}{.53\textwidth}
        \centering
        \begin{adjustbox}{max width=1.00\textwidth}
        \begin{NiceTabular}{c|ccc|c}
        \toprule
        \multirow{2.5}{*}{Dataset} & \multicolumn{3}{c}{SOR-Mamba-SL} & \multirow{2.5}{*}{S-Mamba} \\
        \cmidrule(lr){2-4}
        & Cosine & $\ell_1$ Loss& $\ell_2$ Loss  & \\
        \toprule
        ETTh1 & \first{.442} & \first{.442} & \first{.442} & \second{.457} \\
        ETTh2 & \first{.382} & \first{.382} & \first{.382} & \second{.383} \\
        ETTm1 & \first{.396} & \first{.396} & \first{.396} & \second{.398} \\
        ETTm2 & \first{.284} & \first{.284} & \first{.284} & \second{.290} \\
        PEMS03 & {.145} & .147 & \second{.137} & \first{.133} \\
        PEMS04 & \second{.105} & \second{.105} & {.107} & \first{.096}\\
        PEMS07 & \second{.091} & \second{.091} & \second{.091} & \first{.090}\\
        PEMS08 & {.162} & \second{.159}& {.162} & \first{.157}\\
        Exchange & {.365} & {.365} & \first{.363} & \second{.364}\\
        Weather & \second{.256} & {.257} & {.257} & \first{.252}\\
        Solar & \first{.242} & \first{.242}  & \first{.242} & \second{.244} \\
        ECL & \first{.167} & \second{.168} & .169 & .174\\
        Traffic & \second{.414} & \second{.414} & \first{.412} & .417\\
        \midrule
        \rowcolor{gray!20} Average & \first{.265} & \first{.265}  & \first{.265} & \second{.266}\\
        \bottomrule
        \end{NiceTabular}
        \end{adjustbox}
        \captionsetup{type=table}
        \caption{Robustness to $d$ for regularization.}
        \label{tbl:robust_distance_reg}
    \end{minipage}
    \hfill
    \begin{minipage}{.44\textwidth}
        \centering
        \begin{adjustbox}{max width=1.00\textwidth}
        \begin{NiceTabular}{c|cc|c}
        \toprule
        \multirow{2.5}{*}{Dataset} & \multicolumn{2}{c}{SOR-Mamba-SSL} & \multirow{2.5}{*}{S-Mamba} \\
        \cmidrule(lr){2-3} 
         & $\ell_1$ Loss& $\ell_2$ Loss  & \\
        \toprule
        ETTh1 & \second{.434} & \first{.433} & .457 \\
        ETTh2 & \second{.379} & \first{.376} & .383 \\
        ETTm1 & \first{.391} & \first{.391} & \second{.398} \\
        ETTm2 & \first{.281} & \first{.281} & \second{.290}\\
        PEMS03 & \first{.121} & \first{.121} & \second{.133} \\
        PEMS04 & \second{.099} & \second{.099} & \first{.096}\\
        PEMS07 & \second{.089} & \first{.088} & .090 \\
        PEMS08 & \first{.140} & \second{.142} & .157 \\
        Exchange & \first{.358} & \first{.358} & \second{.364}\\
        Weather & \second{.256} & \second{.256} & \first{.252}\\
        Solar & \second{.232} & \first{.230} & .244 \\
        ECL & \first{.167} & \second{.168} & 174 \\
        Traffic & \first{.402} & \first{.402} & \second{.417} \\
        \midrule
        \rowcolor{gray!20} Average & \second{.258} & \first{.257} & .266\\
        \bottomrule
        \end{NiceTabular}
        \end{adjustbox}
        \captionsetup{type=table}
        \caption{Robustness to $d$ for CCM.}
        \label{tbl:robust_distance_ccm}
    \end{minipage}
\end{figure*}

\vspace{20pt}

\section{Comparison of GPU Memory Usage}
\label{sec:gpu}
\begin{figure*}[h]
\vspace{15pt}
\centering
\includegraphics[width=.83\textwidth]{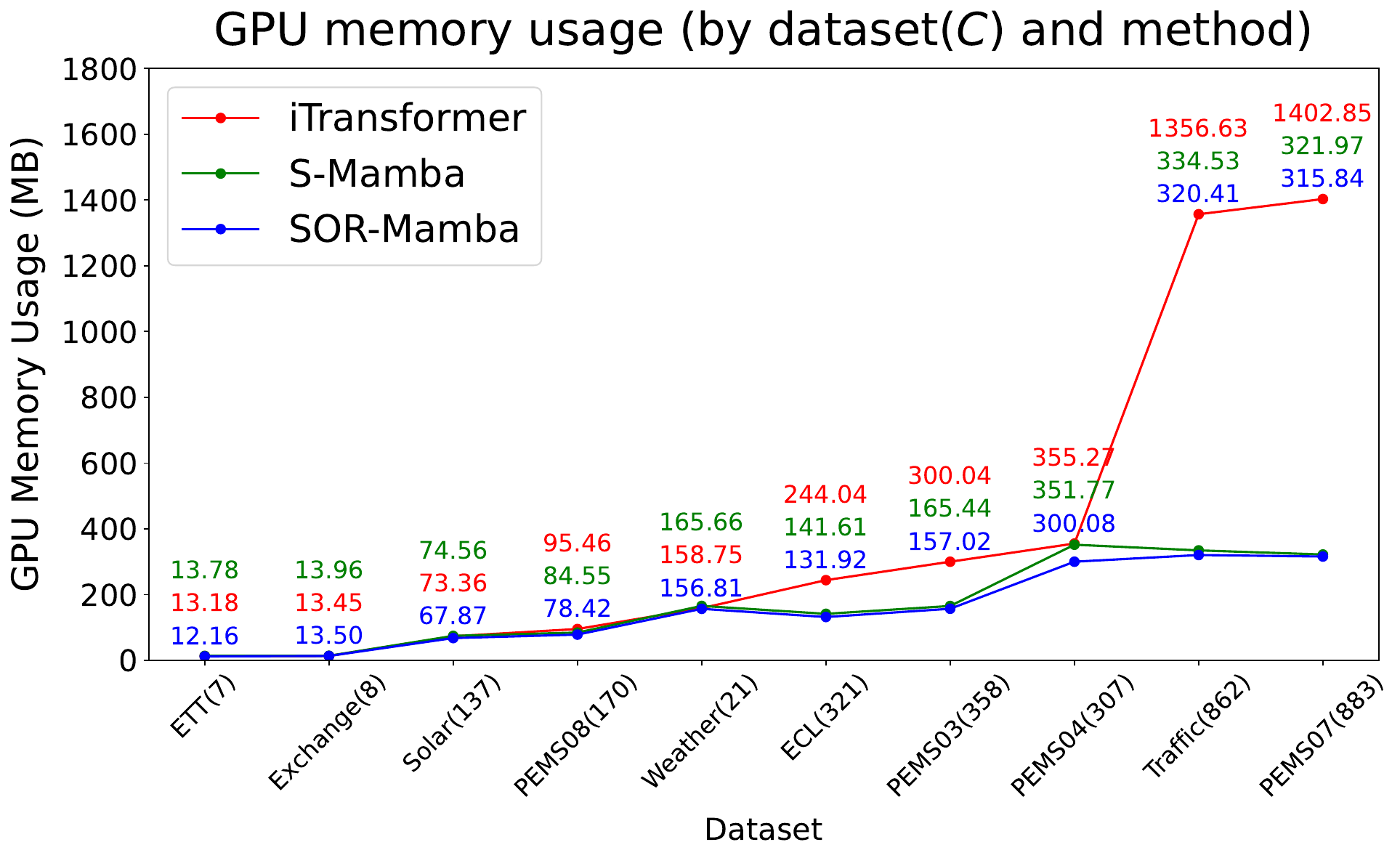} 
\caption{Comparison of GPU memory usage.}
\label{fig:gpu}
\end{figure*}

Figure~\ref{fig:gpu} visualizes GPU memory usage by dataset and method, demonstrating that our method is more efficient than both S-Mamba~\citep{wang2024mamba} and iTransformer~\citep{liu2023itransformer}. Specifically, Mamba-based methods are more efficient than Transformer-based methods when $C$ is large, as Mamba has nearly-linear complexity, whereas Transformers have quadratic complexity.

\newpage
\section{Statistics of Results over Multiple Runs}
To assess the consistency of SOR-Mamba's performance, we present the statistics from results using five different random seeds.
We calculate the mean and standard deviation for both MSE and MAE, detailed in Tables~\ref{tbl:seed5_1}, \ref{tbl:seed5_2}, and \ref{tbl:seed5_3}.
which reveal that our method maintains consistent performance in both self-supervised and supervised settings.

\begin{table*}[h]
\vspace{20pt}
\centering
\begin{adjustbox}{max width=0.999\textwidth}
\begin{NiceTabular}{c|c|cccc}
\toprule
\multicolumn{2}{c}{\multirow{2.5}{*}{Models}} & \multicolumn{4}{c}{Ours} \\
\cmidrule(lr){3-6}
\multicolumn{2}{c}{}&\multicolumn{2}{c}{FT}&\multicolumn{2}{c}{SL}\\
\cmidrule(lr){1-2}\cmidrule(lr){3-4}\cmidrule(lr){5-6}
\multicolumn{2}{c}{Metric} & MSE & MAE & MSE & MAE   \\
\toprule

\multirow{5.5}{*}{\rotatebox{90}{ETTh1}} 
&  96 & $.377_{\pm .001}$&$.398_{\pm .001}$&$.385_{\pm .000}$&$.398_{\pm .000}$\\
& 192 & $.428_{\pm .001}$&$.429_{\pm .000}$&$.432_{\pm .001}$&$.428_{\pm .000}$\\
& 336 & $.464_{\pm .001}$&$.448_{\pm .001}$&$.476_{\pm .000}$&$.448_{\pm .000}$\\
& 720 & $.464_{\pm .001}$&$.469_{\pm .006}$&$.476_{\pm .003}$&$.476_{\pm .002}$\\
\cmidrule(lr){2-6}
& Avg. & $.433_{\pm .000}$&$.436_{\pm .002}$&$.442_{\pm .001}$&$.438_{\pm .000}$\\
\midrule

\multirow{5.5}{*}{\rotatebox{90}{ETTh2}} 
& 96  & $.292_{\pm .004}$&$.348_{\pm .003}$&$.299_{\pm .001}$& $.348_{\pm .001}$\\
& 192 & $.372_{\pm .001}$&$.397_{\pm .001}$&$.375_{\pm .001}$&$.399_{\pm .001}$ \\
& 336 & $.415_{\pm .001}$&$.431_{\pm .000}$&$.423_{\pm .000}$&$.435_{\pm .000}$\\
& 720 & $.423_{\pm .001}$&$.445_{\pm .001}$&$.431_{\pm .002}$&$.446_{\pm .001}$ \\
\cmidrule(lr){2-6}
& Avg. & $.376_{\pm .001}$&$.405_{\pm .001}$&$.382_{\pm .001}$&$.407_{\pm .000}$ \\
\midrule

\multirow{5.5}{*}{\rotatebox{90}{ETTm1}} 
&  96 & $.324_{\pm .002}$&$.362_{\pm .002}$&$.324_{\pm .004}$& $.367_{\pm .003}$\\
& 192 & $.369_{\pm .002}$&$.385_{\pm .001}$&$.375_{\pm .002}$&$.387_{\pm .001}$\\
& 336 & $.402_{\pm .002}$&$.408_{\pm .001}$&$.408_{\pm .000}$&$.408_{\pm .000}$\\
& 720 & $.467_{\pm .002}$&$.444_{\pm .001}$&$.472_{\pm .001}$&$.444_{\pm .001}$ \\
\cmidrule(lr){2-6}
& Avg. & $.391_{\pm .001}$&$.400_{\pm .001}$&$.396_{\pm .001}$&$.401_{\pm .001}$ \\
\midrule

\multirow{5.5}{*}{\rotatebox{90}{ETTm2}} 
& 96  & $.179_{\pm .001}$&$.261_{\pm .001}$&$.181_{\pm .000}$&$.265_{\pm .000}$ \\
& 192 & $.241_{\pm .000}$&$.304_{\pm .000}$&$.246_{\pm .001}$&$.307_{\pm .001}$ \\
& 336 & $.302_{\pm .002}$&$.342_{\pm .002}$&$.306_{\pm .001}$&$.345_{\pm .000}$ \\
& 720 & $.401_{\pm .002}$&$.400_{\pm .002}$&$.403_{\pm .002}$& $.401_{\pm .001}$\\
\cmidrule(lr){2-6}
& Avg. & $.281_{\pm .001}$&$.327_{\pm .000}$&$.284_{\pm .001}$&$.329_{\pm .000}$\\
\bottomrule
\end{NiceTabular}
\end{adjustbox}
\caption{Results of TS forecasting over five runs - 1) ETT datasets.}
\label{tbl:seed5_1}
\end{table*}

\begin{table*}[h]
\centering
\begin{adjustbox}{max width=0.599\textwidth}
\begin{NiceTabular}{c|c|cccc}
\toprule
\multicolumn{2}{c}{\multirow{2.5}{*}{Models}} & \multicolumn{4}{c}{Ours} \\
\cmidrule(lr){3-6}
\multicolumn{2}{c}{}&\multicolumn{2}{c}{FT}&\multicolumn{2}{c}{SL}\\
\cmidrule(lr){1-2}\cmidrule(lr){3-4}\cmidrule(lr){5-6}
\multicolumn{2}{c}{Metric} & MSE & MAE & MSE & MAE   \\
\toprule
\multirow{5.5}{*}{\rotatebox{90}{PEMS03}} 
& 12&$.066_{\pm .001}$&$.170_{\pm .001}$&$.066_{\pm .001}$&$.170_{\pm .001}$\\
& 24&$.088_{\pm .001}$&$.197_{\pm .001}$&$.090_{\pm .001}$&$.200_{\pm .001}$\\
& 48 & $.134_{\pm .002}$&$.245_{\pm .003}$&$.167_{\pm .001}$&$.280_{\pm .001}$ \\
& 96 & $.193_{\pm .005}$&$.297_{\pm .006}$&$.225_{\pm .003}$&$.318_{\pm .002}$ \\
\cmidrule(lr){2-6}
& Avg. & $.121_{\pm .002}$&$.227_{\pm .002}$&$.137_{\pm .001}$&$.242_{\pm .001}$ \\
\midrule

\multirow{5.5}{*}{\rotatebox{90}{PEMS04}} 
& 12 & $.074_{\pm .002}$&$.175_{\pm .003}$&$.077_{\pm .000}$&$.180_{\pm .000}$ \\
& 24 & $.086_{\pm .003}$&$.192_{\pm .005}$&$.091_{\pm .001}$&$.197_{\pm .001}$ \\
& 48 & $.106_{\pm .001}$&$.214_{\pm .005}$&$.115_{\pm .002}$&$.221_{\pm .003}$ \\
& 96 & $.129_{\pm .003}$&$.233_{\pm .004}$&$.143_{\pm .002}$&$.248_{\pm .002}$ \\
\cmidrule(lr){2-6}
& Avg. & $.099_{\pm .001}$&$.203_{\pm .002}$&$.107_{\pm .001}$&$.212_{\pm .001}$ \\
\midrule

\multirow{5.5}{*}{\rotatebox{90}{PEMS07}} 
& 12&$.059_{\pm .001}$&$.155_{\pm .001}$&$.060_{\pm .000}$&$.156_{\pm .000}$\\
& 24&$.076_{\pm .005}$&$.174_{\pm .004}$&$.082_{\pm .000}$&$.182_{\pm .000}$\\
& 48&$.098_{\pm .001}$&$.199_{\pm .001}$&$.107_{\pm .001}$&$.209_{\pm .000}$ \\
&96&$.117_{\pm .003}$&$.218_{\pm .003}$&$.117_{\pm .001}$&$.218_{\pm .001}$\\
\cmidrule(lr){2-6}
& Avg.&$.088_{\pm .001}$&$.186_{\pm .001}$&$.091_{\pm .000}$&$.191_{\pm .000}$\\
\midrule

\multirow{5.5}{*}{\rotatebox{90}{PEMS08}}
& 12& $.078_{\pm .000}$&$.178_{\pm .000}$&$.076_{\pm .001}$&$.176_{\pm .000}$\\
& 24&$.103_{\pm .001}$&$.205_{\pm .002}$&$.109_{\pm .001}$&$.212_{\pm .001}$\\
& 48&$.159_{\pm .001}$&$.250_{\pm .001}$&$.172_{\pm .003}$&$.264_{\pm .003}$\\
& 96 & $.229_{\pm .001}$&$.295_{\pm .002}$&$.290_{\pm .002}$&$.334_{\pm .002}$ \\
\cmidrule(lr){2-6}
& Avg.&$.142_{\pm .000}$&$.232_{\pm .001}$&$.162_{\pm .001}$&$.247_{\pm .001}$\\
\bottomrule
\end{NiceTabular}
\end{adjustbox}
\caption{Results of TS forecasting over five runs - 2) PEMS datasets.}
\label{tbl:seed5_2}
\end{table*}

\begin{table*}[h]
\centering
\begin{adjustbox}{max width=0.599\textwidth}
\begin{NiceTabular}{c|c|cccc}
\toprule
\multicolumn{2}{c}{\multirow{2.5}{*}{Models}} & \multicolumn{4}{c}{Ours} \\
\cmidrule(lr){3-6}
\multicolumn{2}{c}{}&\multicolumn{2}{c}{FT}&\multicolumn{2}{c}{SL}\\
\cmidrule(lr){1-2}\cmidrule(lr){3-4}\cmidrule(lr){5-6}
\multicolumn{2}{c}{Metric} & MSE & MAE & MSE & MAE   \\
\toprule
\multirow{5.5}{*}{\rotatebox{90}{Exchange}} 
& 96  & $.085_{\pm .001}$&$.204_{\pm .002}$&$.085_{\pm .001}$&$.205_{\pm .001}$\\
& 192 & $.179_{\pm .000}$&$.301_{\pm .000}$&$.179_{\pm .002}$&$.301_{\pm .001}$\\

& 336 & $.329_{\pm .001}$& $.415_{\pm .001}$&$.331_{\pm .000}$&$.417_{\pm .000}$\\

& 720 & $.838_{\pm .005}$&$.690_{\pm .002}$&$.860_{\pm .001}$&$.698_{\pm .001}$\\
\cmidrule(lr){2-6}

& Avg. & $.358_{\pm .001}$&$.402_{\pm .001}$&$.363_{\pm .001}$&$.405_{\pm .001}$\\
\midrule

\multirow{5.5}{*}{\rotatebox{90}{Weather}} 
& 96  & $.174_{\pm .000}$&$.212_{\pm .000}$&$.175_{\pm .001}$&$.215_{\pm .000}$ \\

& 192 & $.221_{\pm .000}$&$.255_{\pm .000}$&$.221_{\pm .000}$&$.255_{\pm .000}$\\

& 336 & $.277_{\pm .000}$&$.295_{\pm .001}$&$.277_{\pm .001}$&$.296_{\pm .001}$\\

& 720 & $.353_{\pm .001}$&$.348_{\pm .001}$&$.355_{\pm .000}$& $.348_{\pm .000}$\\
\cmidrule(lr){2-6}

& Avg. & $.256_{\pm .000}$&$.277_{\pm .000}$&$.257_{\pm .000}$& $.278_{\pm .000}$\\
\midrule

\multirow{5.5}{*}{\rotatebox{90}{Solar}} 
& 96  & $.194_{\pm .005}$&$.229_{\pm .004}$&$.207_{\pm .000}$&$.246_{\pm .001}$ \\

& 192 & $.228_{\pm .002}$&$.256_{\pm .003}$&$.239_{\pm .001}$&$.270_{\pm .001}$\\

& 336 & $.247_{\pm .006}$&$.276_{\pm .005}$&$.260_{\pm .001}$&$.287_{\pm .001}$\\

& 720 & $.251_{\pm .003}$&$.275_{\pm .003}$&$.264_{\pm .001}$&$.291_{\pm .001}$\\
\cmidrule(lr){2-6}

& Avg. & $.230_{\pm .002}$&$.259_{\pm .002}$&$.242_{\pm .000}$&$.274_{\pm .000}$\\
\midrule

\multirow{5.5}{*}{\rotatebox{90}{ECL}} 
& 96  & $.139_{\pm .001}$&$.235_{\pm .002}$&$.139_{\pm .001}$&$.233_{\pm .001}$\\

& 192 & $.160_{\pm .002}$&$.254_{\pm .002}$&$.158_{\pm .001}$&$.249_{\pm .001}$\\

& 336 & $.176_{\pm .003}$&$.271_{\pm .003}$&$.177_{\pm .001}$&$.271_{\pm .001}$\\
& 720 & $.198_{\pm .003}$&$.292_{\pm .006}$&$.201_{\pm .003}$&$.293_{\pm .002}$\\
\cmidrule(lr){2-6}
& Avg. & $.168_{\pm .001}$&$.264_{\pm .001}$&$.169_{\pm .001}$&$.262_{\pm .001}$\\
\midrule

\multirow{5.5}{*}{\rotatebox{90}{Traffic}} 
&96&$.378_{\pm .000}$&$.258_{\pm .000}$&$.378_{\pm .000}$&$.259_{\pm .000}$\\
& 192&$.393_{\pm .001}$&$.267_{\pm .001}$&$.399_{\pm .000}$&$.270_{\pm .000}$\\
& 336&$.399_{\pm .001}$&$.276_{\pm .002}$&$.416_{\pm .001}$&$.279_{\pm .000}$\\
& 720&$.437_{\pm .001}$&$.289_{\pm .002}$&$.456_{\pm .001}$&$.297_{\pm .001}$\\
\cmidrule(lr){2-6}
& Avg.&$.402_{\pm .000}$&$.273_{\pm .001}$&$.412_{\pm .000}$&$.276_{\pm .000}$\\
\bottomrule
\end{NiceTabular}
\end{adjustbox}
\caption{Results of TS forecasting over five runs - 3) Other datasets.}
\label{tbl:seed5_3}
\end{table*}

\end{document}